\documentclass[11pt,a4paper]{article}
\usepackage{times,latexsym}
\usepackage{url}
\usepackage{graphicx}
\usepackage[T1]{fontenc}
\usepackage{booktabs}
\usepackage{color}
\usepackage{xcolor}
\usepackage{colortbl}
\usepackage{multirow}
\usepackage[]{caption}
\usepackage{subcaption}
\usepackage{framed}
\usepackage{soul}
\usepackage[export]{adjustbox} %
\usepackage{float}
\usepackage{amsmath}

\newcommand{\Hrule}[3][.]{%
  \par\addvspace{#2}%
  \begingroup\color{#1}%
  \hrule
  \endgroup
  \addvspace{#3}%
}

\newcommand{\slashn}{\textbackslash n}

\newenvironment{cframed}[1][gray!40]
  {\def\FrameCommand{\fboxsep=\FrameSep\fcolorbox{#1}{white}}%
    \MakeFramed {\advance\hsize-\width \FrameRestore}}
  {\endMakeFramed}

\newcommand\nl[1]{\textit{#1}}

\definecolor{forestgreen}{rgb}{0.13, 0.55, 0.13}

\definecolor{grey}{gray}{0.9}

\usepackage[acceptedWithA]{tacl2021v1}

\usepackage{xspace,mfirstuc,tabulary}

\newif\iftaclinstructions
\taclinstructionsfalse %
\iftaclinstructions
\renewcommand{\confidential}{}
\renewcommand{\anonsubtext}{(No author info supplied here, for consistency with
TACL-submission anonymization requirements)}
\newcommand{\instr}
\fi

\iftaclpubformat %

\else

\fi

\title{Robust Pronoun Fidelity with English LLMs: \\
Are they Reasoning, Repeating, or Just Biased?}

\author{
  Vagrant Gautam\textsuperscript{1} \,
  Eileen Bingert\textsuperscript{1} \,
  Dawei Zhu\textsuperscript{1} \,
  \textbf{Anne Lauscher\textsuperscript{2}} \,
  \textbf{Dietrich Klakow\textsuperscript{1}}
  \\
  \textsuperscript{1}Saarland University, Germany \\
  \textsuperscript{2}Data Science Group, University of Hamburg, Germany
  \\
  \texttt{vgautam@lsv.uni-saarland.de}
  \\
}

\date{}

\begin{document}
\maketitle
\begin{abstract}
Robust, faithful and harm-free pronoun use for individuals is an important goal for language model development as their use increases, but prior work tends to study only one or two of these characteristics at a time.
To measure progress towards the combined goal, we introduce the task of \textit{pronoun fidelity}: given a context introducing a co-referring entity and pronoun, the task is to reuse the correct pronoun later.
We present \textsc{RUFF}, a carefully-designed dataset of over 5 million instances to measure robust pronoun fidelity in English, and we evaluate 37 model variants from nine popular families, across architectures (encoder-only, decoder-only and encoder-decoder) and scales (11M-70B parameters).
When an individual is introduced with a pronoun, models can mostly faithfully reuse this pronoun in the next sentence, but they are significantly worse with \textit{she/her/her}, singular \textit{they} and neopronouns.
Moreover, models are easily distracted by non-adversarial sentences discussing other people;
even one sentence with a distractor pronoun causes accuracy to drop on average by 34 percentage points.
Our results show that pronoun fidelity is not robust, in a simple, naturalistic setting where humans achieve nearly 100\% accuracy.
We encourage researchers to bridge the gaps we find and to carefully evaluate reasoning in settings where superficial repetition might inflate perceptions of model performance.\looseness=-1
\end{abstract}

\section{Introduction}

Third-person pronouns (\emph{he}, \emph{she}, \emph{they}, etc.) are words that construct individuals' identities in conversations~\citep{SILVERSTEIN1985219}.
In English, these pronouns mark referential gender for the entity they are referring to,
which can also index an individual's social gender, e.g., man, woman, non-binary~\citep{cao-daume-iii-2020-toward}.
Correctly using the pronouns an individual identifies with is important, as misgendering (including through incorrect pronoun use)
can in the best case be a social faux pas~\citep{stryker2017transgender} and in the worst case, cause psychological distress, particularly to transgender individuals~\citep{McLemore_2018}.

Accordingly, it is important for large language models (LLMs) to use pronouns faithfully and without causing harm.
To this end, many studies have explored how LLMs handle pronouns, showing that they stereotypically associate pronouns and occupations~\citep{kurita-etal-2019-measuring}, reason about co-referring pronouns and entities better when they conform to stereotypes~\citep{tal-etal-2022-fewer}, fail when exposed to novel pronoun phenomena such as neopronouns~\citep{lauscher-etal-2023-em}, and cannot consistently reuse neopronouns during generation~\citep{ovalle23TANGO}.
These shortcomings create differences in quality of service and cause representational harm,
amplifying discrimination against certain pronoun users~\citep{blodgett-etal-2020-language,dev-etal-2021-harms}.

In work on LLM pronoun use, a question that has gone unexamined thus far is: \textbf{\emph{How robust is model faithfulness to pronouns}} when discussing more than one person? To answer this question, we propose \emph{pronoun fidelity} (\S\ref{sec:task-definition}), a new task to investigate realistic model reasoning about pronouns, and we introduce \textsc{RUFF} (\S\ref{sec:ruff-dataset}), a novel, large-scale dataset of over 5 million instances, to evaluate this task. With this dataset, we present an analysis of pronoun fidelity across 37 variants from nine popular language model families covering architectures and scales, to investigate whether models are reasoning, repeating, or just biased.

\begin{figure}[t!]
    \centering
    \includegraphics[valign=m,width=\linewidth]{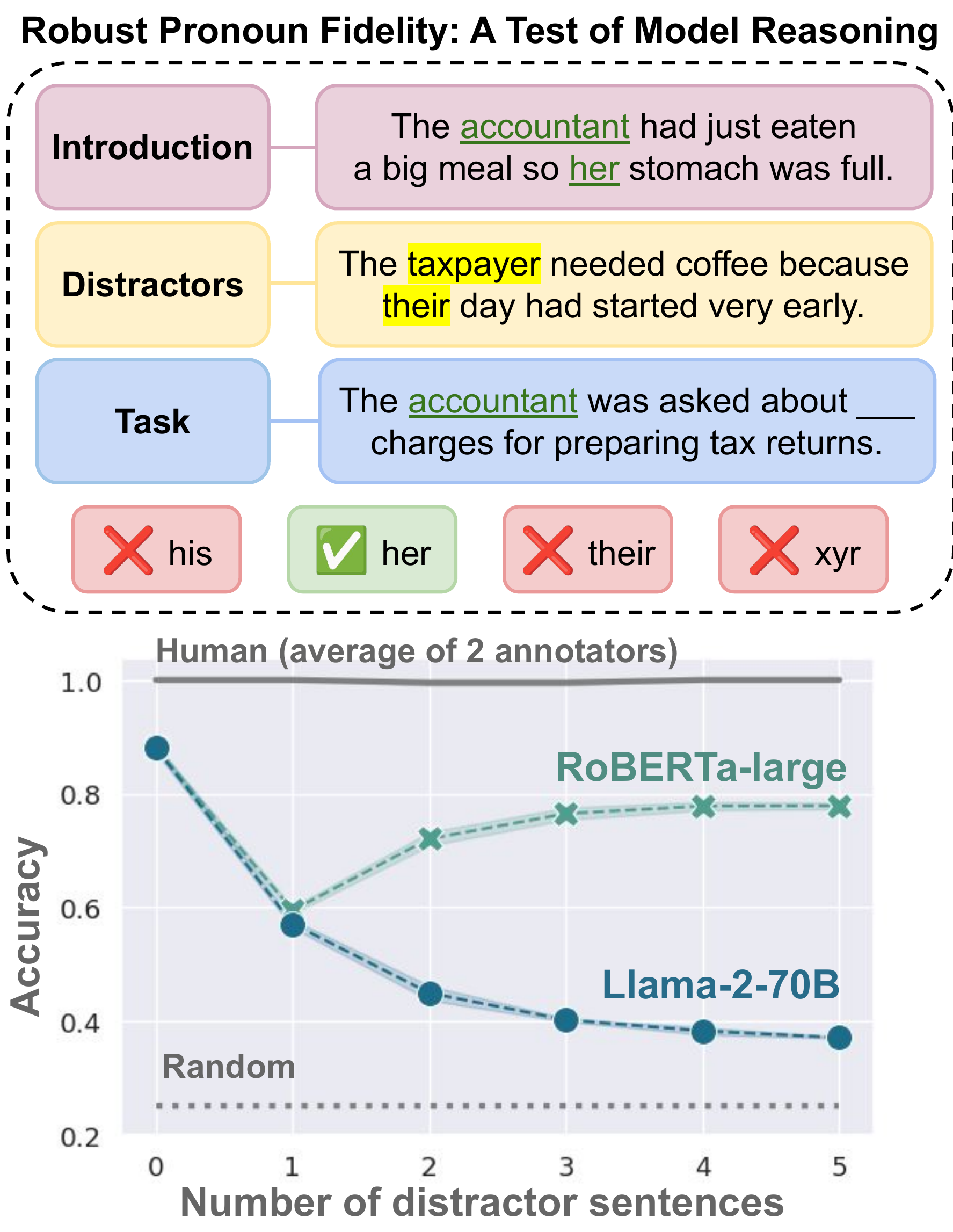}
    \caption{We evaluate model accuracy at using the correct pronoun for an entity when provided with an explicit introduction and 0-5 non-adversarial distractor sentences. \textsc{Llama-2-70B} and \textsc{RoBERTa-large} show large accuracy drops with just one distractor. Accuracy is averaged over 3 data splits; standard deviation is shown with shading.}
    \label{fig:one}
    \vspace{-3mm}
\end{figure}

First, we collect model pronoun predictions for occupations in the absence of context, to establish a ``bias baseline'' (\S\ref{sec:context-free-model-predictions}).
Next, we evaluate whether models can overcome their biased pronoun predictions when explicitly shown what pronoun to use in context (\S\ref{sec:injecting-an-introductory-context}).
All models are good at this task, but there are significant disparities across pronoun sets.
We then test the robustness of this result by inserting naturalistic distractor sentences using a different pronoun to talk about another person (\S\ref{sec:injecting-distractors}).
\textit{Even one non-adversarial distractor sentence vastly deteriorates model performance} as shown in Figure \ref{fig:one}.
Finally, in a detailed error analysis (\S\ref{error-analysis}), we disentangle whether model errors can be attributed to distraction or falling back to bias, finding that encoder-only and decoder-only models behave in fundamentally different ways.

Overall, our results show that \textit{models struggle to reason about pronouns in a simple, naturalistic setting} and highlight the need for careful task design to ensure that superficial repetition does not lead to inflated claims about model reasoning.
We release all code and data to encourage researchers to bridge the gaps we find:
\url{https://github.com/uds-lsv/robust-pronoun-fidelity}

\section{Pronoun Fidelity Task}
\label{sec:task-definition}
Discussing multiple individuals is natural, frequent and well-studied in discourse;
we use both definite references and pronouns in natural language to establish continuity and coherence~\citep{grosz-etal-1995-centering}.
We formalize a version of these phenomena in our task:
given a context in which a co-referring entity and pronoun are introduced, the task is to reconstruct the pronoun later in a sentence about the entity, independent of a limited number of potential distractors.

\begin{cframed}
\noindent\textbf{Introduction:} \nl{The \textcolor{forestgreen}{\underline{accountant}} had just eaten a big meal so \textcolor{forestgreen}{\underline{her}} stomach was full.}

\vspace{0.05in}
\Hrule[gray!40]{3pt}{5pt}
\vspace{0.05in}

\centerline{\textsc{(Optional)}}
\noindent\textbf{Distractor 1:} \nl{The \hl{taxpayer} needed coffee because \hl{their} day had started very early.}
\centerline{\dots}
\noindent\textbf{Distractor N:} \nl{\hl{Their} sleep had been fitful.}

\vspace{0.05in}
\Hrule[gray!40]{3pt}{5pt}
\vspace{0.05in}

\noindent\textbf{Task sentence:} \nl{The \textcolor{forestgreen}{\underline{accountant}} was asked about \_\_\_ charges for preparing tax returns.}

\end{cframed}

More formally, an introduction sentence $i(\textcolor{forestgreen}{e_a}, \textcolor{forestgreen}{p_a})$ establishes a coreference between an entity \textcolor{forestgreen}{$e_a$} and a pronoun \textcolor{forestgreen}{$p_a$}. A distractor sentence $d(\text{\hl{$e_b$}}, \text{\hl{$p_b$}} )$ explicitly establishes or implicitly continues a previously-established coreference between a different entity \hl{$e_b$} and a different pronoun \hl{$p_b$}, i.e., \textcolor{forestgreen}{$e_a$}$ \neq $\hl{${e_b}$} and \textcolor{forestgreen}{$p_a$}$ \neq $\hl{$p_b$}. Let $\mathcal{D}(\text{\hl{$e_b$}}, \text{\hl{$p_b$}})$ be a set of distractor sentences such that $0 \leq |\mathcal{D}(\text{\hl{$e_b$}}, \text{\hl{$p_b$}})| \leq N$. When combined, an introduction sentence and the set of distractor sentences form a context. A task sentence $t(\textcolor{forestgreen}{e_a}, p)$ contains an unambiguous coreference between the entity \textcolor{forestgreen}{$e_a$} from the introduction and a pronoun slot $p$ which must be filled.
The task is to maximize
\vspace{-0.05cm}
\begin{equation}
    P[\ t(\textcolor{forestgreen}{e_a}, p=\textcolor{forestgreen}{p_a})\ | \ 
    i(\textcolor{forestgreen}{e_a}, \textcolor{forestgreen}{p_a}),\ 
    \mathcal{D}(\text{\hl{$e_b$}}, \text{\hl{$p_b$}})
    \ ] \,,
\end{equation}
the probability $P$ of reconstructing the correct pronoun \textcolor{forestgreen}{$p_a$} in the sentence $t(\textcolor{forestgreen}{e_a}, p)$, given the context.

\section{\textsc{RUFF} Dataset}
\label{sec:ruff-dataset}

\begin{figure*}
    \centering
    \includegraphics[width=\linewidth]{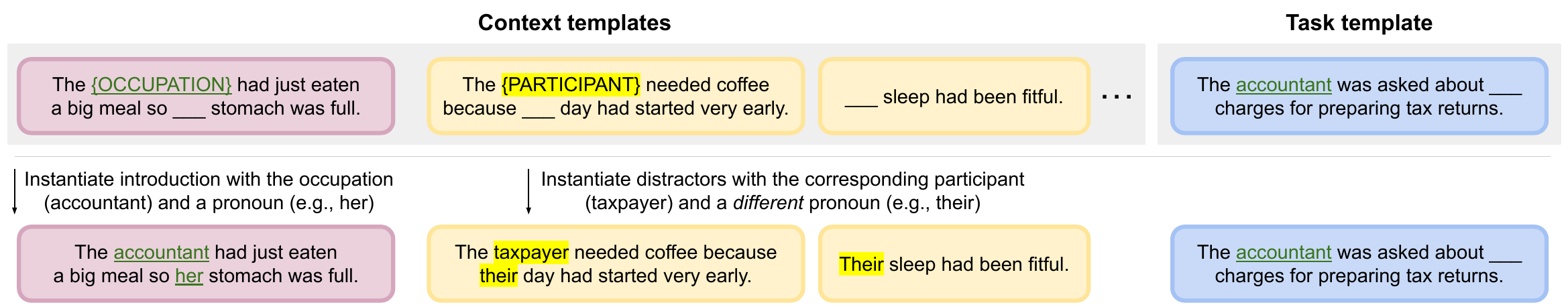}
    \caption{Template assembly for \textsc{RUFF}: occupation-specific task templates are matched with generic context templates (introductions and optional distractors) that are instantiated with disjoint pronoun sets. This creates realistic but controlled narratives that allow us to measure robust pronoun fidelity.}
    \label{fig:template-assembly}
\end{figure*}

To evaluate \textbf{R}obust prono\textbf{U}n \textbf{F}idelity at scale, we create \textsc{RUFF}, an evaluation dataset of narratives.
Each dataset instance describes a simple narrative with 1-2 people, but rather than narrative data that focuses on commonsense event reasoning~\citep{mostafazadeh-etal-2016-corpus}, we focus on pronominal reasoning, like \citet{rudinger-etal-2018-gender}.

\noindent Specifically, we examine four third-person pronouns in three grammatical cases (nominative, accusative and possessive dependent);
in addition to the English masculine (\textit{he/him/his}) and feminine (\textit{she/her/her}) pronouns, we heed \citeauthor{lauscher-etal-2022-welcome}'s \citeyearpar{lauscher-etal-2022-welcome} call for more inclusive NLP research by examining two more pronoun sets that are less well-studied in NLP: singular \textit{they} (\textit{they/them/their}), the pronoun of choice of over 75\% of respondents to the Gender Census~\citep{Lodge_2023}, and \textit{xe/xem/xyr}, the most popular neopronoun according to the same census.
Our narratives cover 60 occupations and corresponding participants (see Appendix~\ref{sec:occupation-list}), following Winogender schemas~\citep{rudinger-etal-2018-gender}, as their bias characteristics are well-studied in NLP.
In total, \textsc{RUFF} contains over 5 million data instances.
Each instance is designed to have an unambiguous answer, and is constructed with a 3-step pipeline: template creation (\S\ref{sec:template-creation}), template assembly (\S\ref{sec:template-assembly}), and data validation (\S\ref{sec:data-validation}).

\subsection{Template creation}
\label{sec:template-creation}

Below, we describe how we create occupation-specific task templates and generic context templates for introductions and distractors.\\

\noindent\textbf{Task templates.}\quad
We create one task sentence template per occupation and grammatical case, with an unambiguous, unique coreference between the pronoun and occupation, for a total of 180 templates.
For instance, \textit{charges for preparing tax returns} can only belong to an \textit{accountant}, never a \textit{taxpayer}, the corresponding participant.\\

\noindent\textbf{Context templates.}\quad
The ideal context template would be: \textbf{(1)} \textbf{flexible} across different occupations and participants, for a controlled setting to test robustness; \textbf{(2)} \textbf{cohesive} in a multi-sentence narrative leading up to the task template about an occupation; and \textbf{(3)} \textbf{neutral}, not dramatically affecting the prediction of a certain pronoun.
Templates such as \textit{He is an accountant} are well-established for testing word embedding associations~\citep{caliskan2016semantics,may-etal-2019-measuring}.
They are flexible and neutral (they are even referred to as ``semantically bleached'' templates in the literature), but it is unnatural to use more than one consecutively.
Natural corpora like \citet{levy-etal-2021-collecting-large} have the most potential for creating cohesive narratives, but contain occupation-specific sentences that are inflexible and sometimes also non-neutral, e.g., ungrammatical with singular \textit{they}.

For a setting that satisfies all three criteria, we create context templates with generic themes, e.g., universal human emotions and sensations (\textit{hungry}/\textit{full}, \textit{tired}/\textit{energetic}, \textit{unhappy}/\textit{happy}, etc.).
The generic themes make them flexible for use across all occupations and participants.
Templates of the same polarity can be stacked into a cohesive narrative, e.g., a narrative about a taxpayer having a bad day after sleeping poorly and missing a meal.
Our templates are created to be grammatical with all pronoun sets we consider, which satisfies neutrality.
Additionally, our use of both positive and negative versions of templates (i.e., \textit{happy} and \textit{unhapppy}) as well as our variety of templates allows us to mitigate potential implicit biases when aggregated~\citep{alnegheimish-etal-2022-using}.

To reflect natural and coherent use of pronouns in discourse, we create explicit (definite reference + pronoun) and implicit (pronoun-only) versions of 10 context templates per grammatical case, for a total of 30 templates.
Each explicit context template begins with an entity and introduces the pronoun in a clause, e.g., \textit{The taxpayer needed coffee because their day had started very early}, while implicit templates are simple sentences like \textit{Their sleep had been fitful}.
See Appendix~\ref{sec:context-template-details} for more detail on template creation and assembly.

\subsection{Template assembly}
\label{sec:template-assembly}

Figure \ref{fig:template-assembly} shows how we instantiate and combine templates to assemble our data instances: first, we select an occupation (\textcolor{forestgreen}{$e_a$}) and one of its task templates. We pick a pronoun (\textcolor{forestgreen}{$p_a$}) to use as ground truth and instantiate a random context template with the selected occupation and pronoun.
The simplest version of the pronoun fidelity task includes just this introduction sentence followed by the task sentence. Instantiating 10 templates with 4 different pronoun sets and pairing them with task templates for 60 occupations across 3 grammatical cases gives us a total of 7,200 unique instances for this version of the task.

To create more complex data instances, we insert a variable number of distractor sentences between the introduction and task sentences, discussing a participant \hl{$e_b$} with a \textit{different} pronoun \hl{$p_b$}.
These are also sampled from the set of context templates (see Appendix~\ref{sec:context-template-details} for details).
Instantiating 4 templates with 3 previously unused pronouns gives 86,400 unique instances with one distractor.

\begin{table}[t!]
    \centering
        \begin{tabular}{lr}
        \toprule
        \bf Data type\phantom{space} & \bf \phantom{space}Number of instances \\
        \midrule
        \rowcolor{grey} \multicolumn{2}{c}{With no context} \\
        Task sentences & 180 \\
        \midrule
        \rowcolor{grey} \multicolumn{2}{c}{With introductory context} \\
        + 0 distractors & 3 x 2,160 \phantom{0,00}(of 7,200)  \\
        + 1 distractor\phantom{s} & 3 x 2,160 \phantom{0,0}(of 86,400) \\
        + 2 distractors & 3 x 2,160 \phantom{0,}(of 345,600) \\
        + 3 distractors & 3 x 2,160 (of 1,036,800) \\
        + 4 distractors & 3 x 2,160 (of 2,073,600) \\
        + 5 distractors & 3 x 2,160 (of 2,073,600) \\
        \bottomrule
    \end{tabular}
    \caption{Number of dataset instances. Pronoun fidelity instances consist of task instances combined with introductory contexts and optional distractors. We subsample 3 sets of 2,160 sentences (of the total number of instances we created).}
    \vspace{-5mm}
    \label{tab:data-statistics}
\end{table}

Our stackable dataset design allows us to generate a vast amount of data of varying lengths, giving us a controlled setting to evaluate context effects on model predictions.
We subsample the data with three random seeds for the rest of our evaluation, ensuring that all occupations, cases, pronoun declensions and distractor pronouns are equally represented in each subsampled set of 2,160 sentences.
All data statistics are shown in Table \ref{tab:data-statistics}.

\begin{table*}[ht!]
    \centering
    \begin{tabular}{lll}
    \toprule
    \multicolumn{1}{c}{\textbf{Model}} & \multicolumn{1}{c}{\textbf{Sizes}} & \multicolumn{1}{c}{\textbf{Architecture}} \\
    \midrule
    \rowcolor{grey} \multicolumn{3}{c}{Evaluated with (Pseudo) Log Likelihoods} \\
        \textsc{ALBERT-v2} & base (11M), large (17M), xlarge (58M), xxlarge (223M) & Encoder-only \\
        \textsc{BERT} & base (110M), large (340M) & Encoder-only \\
        \textsc{RoBERTa} & base (125M), large (355M) & Encoder-only \\
        \textsc{MosaicBERT} & 137M & Encoder-only \\
        \textsc{OPT} & 125M, 350M, 1.3B, 2.7B, 6.7B, 13B, 30B, 66B & Decoder-only \\
        \textsc{Pythia} & 14M, 70M, 160M, 410M, 1B, 1.4B, 2.8B, 6.9B, 12B  & Decoder-only \\
        \textsc{Llama-2} & 7B, 13B, 70B & Decoder-only \\
         \midrule
         \rowcolor{grey} \multicolumn{3}{c}{Evaluated with prompting} \\
         \textsc{FLAN-T5} & small (77M), base (248M), large (783M), xl (2.85B), xxl (11.3B) & Encoder-decoder \\
         \textsc{Llama-2-chat} & 7B, 13B, 70B & Decoder-only \\
         \bottomrule
    \end{tabular}
    \caption{Models we experiment with across a range of sizes (11M-70B parameters) and architectures.}
    \label{tab:models}
\end{table*}

\subsection{Data validation}
\label{sec:data-validation}

We validate all task and context templates. To verify that the pronoun fidelity task is easy and unambiguous for humans, and to create a ceiling for model evaluation, we also validate a subset of task instances with 0-5 distractors.
Annotator information is shown in Appendix~\ref{sec:annotator-information} and all annotator instructions are provided in Appendix~\ref{sec:annotation-instructions}. \\

\noindent\textbf{Templates.}\quad
Two authors with linguistic training iteratively created and validated sentence templates for grammaticality and correct coreferences until consensus was reached.
An additional annotator independently rated 100\% of the sentences as grammatical and with the correct coreferences. \\

\noindent\textbf{Pronoun fidelity task.}\quad
We sampled 100 instances with each possible number of distractors (0-5), for a total of 600 instances. One author and one annotator had to fill in the pronoun and they each performed with 99.8\% accuracy.\footnote{They disagreed on non-overlapping instances which appeared to be random slips.}

\section{Experimental Setup}

\begin{figure*}
    \centering
    \includegraphics[width=\linewidth]{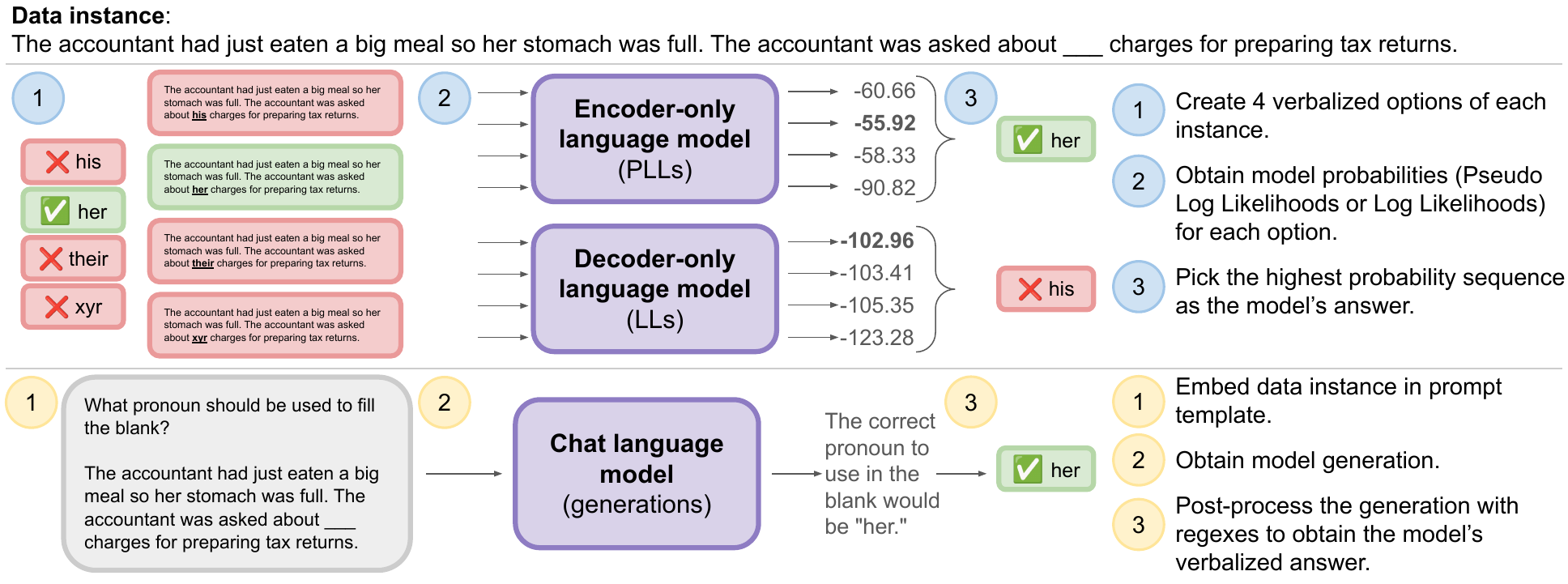}
    \vspace{-8mm}
    \caption{Model evaluation overview: pseudo log likelihoods (PLLs) and log likelihoods (LLs) of verbalized instances are used for encoder-only and decoder-only models; generations are used for chat models.\looseness=-1}
    \label{fig:evaluation}
    \vspace{-4mm}
\end{figure*}

We list our models, evaluation methods, and metrics. Further details are provided in Appendix \ref{sec:experimental-details}.

\subsection{Models}
We experiment with 37 transformer-based language model variants from nine popular model families (see Table \ref{tab:models}), which we chose to evaluate the effects of architecture and scaling.
Our encoder-only models are from the \textsc{BERT}~\citep{devlin-etal-2019-bert}, \textsc{RoBERTa}~\citep{liu2019roberta}, \textsc{ALBERT-v2}~\citep{ALBERT_ICLR} and \textsc{MosaicBERT}~\citep{MosaicBERT_NeurIPS} model families, as the first three remain well-used in NLP, and the last is trained on much more data.
As for our decoder-only models, we select the popular \textsc{Llama-2}~\citep{touvron2023llama} model family, as well as \textsc{OPT}~\citep{zhang2022opt} and \textsc{Pythia}~\citep{OPT_ICML} for their large range of model sizes.
In Appendix \ref{sec:prompting-results}, we also experiment with popular chat models that are further trained with instruction-tuning and reinforcement learning, to evaluate task performance with prompting; specifically, we use decoder-only \textsc{Llama-2-chat} models~\citep{touvron2023llama} and encoder-decoder \textsc{FLAN-T5} models~\citep{chung2022scaling}.

\subsection{Obtaining predictions}

Figure \ref{fig:evaluation} shows an overview of our evaluation methods.
Decoder-only and encoder-only models are evaluated comparably in a forced choice setting: following  \citet{hu-levy-2023-prompting}, we take direct measurements of probabilities as a proxy for models' metalinguistic judgements.
Generations are obtained from chat models and post-processed to obtain unique pronouns, if any.\\

\noindent\textbf{Encoder-only and decoder-only models.}\quad
We verbalize four versions of each data instance, i.e., we fill in the blank with each of the four pronouns we consider, creating four options.
We then obtain model probabilities for each of these four options, and select the highest probability option as the model's choice.
We use log likelihoods for decoder-only models and pseudo log likelihoods for encoder-only models, following prior work~\citep{salazar-etal-2020-masked, kauf-ivanova-2023-better}.
We do not use masked token prediction due to tokenization issues with neopronouns~\citep{ovalle2023talking}; briefly, we want \textit{xe} to be tokenized ``normally'' (which is often as two tokens) rather than a single \texttt{UNK} token. \\

\noindent\textbf{Chat models.}\quad
Following common practice, we evaluate chat models (\textsc{FLAN-T5} and \textsc{Llama-2-chat}) using vanilla and chain-of-thought prompting.
Following~\citet{sclar2024quantifying}, we show the range of expected performance with 10 different prompts, inspired by the prompts to elicit coreferences in the \textsc{FLAN} collection~\citep{flancollection23}.
See Appendix~\ref{sec:prompting} for more methodological details and Appendix~\ref{sec:prompting-results} for results.

\subsection{Metrics}
As every instance of the pronoun fidelity task has a unique correct answer, we report \textit{accuracy} averaged over the three randomly sampled subsets of our dataset. We show the standard deviation with error bars or shading.
Where possible, we perform significance testing with a Welch's t-test and a threshold of 0.05. We use human performance as our ceiling, and compare models to a baseline of randomly selecting 1 of the 4 pronouns (i.e., 25\%).

\begin{figure}[t!]
    \centering
    \includegraphics[width=\linewidth]{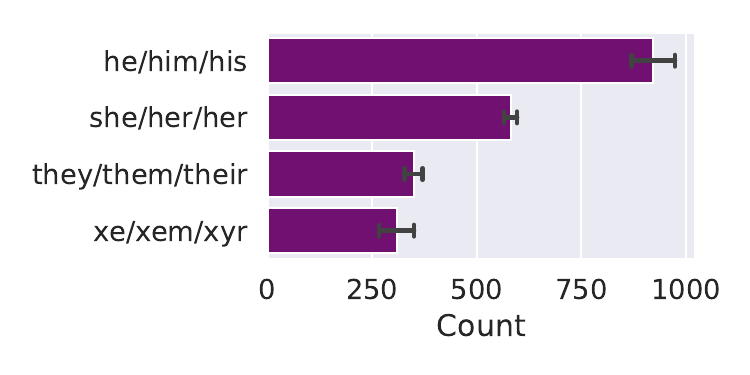}
    \vspace{-10mm}
    \caption{Counts of pronoun predictions from all models, in the absence of context. Error bars indicate standard deviation across models.}
    \label{fig:context-free-model-predictions}
    \vspace{-4mm}
\end{figure}

\section{Model Predictions with No Context}
\label{sec:context-free-model-predictions}

We begin by creating a ``bias baseline,'' i.e., obtaining pronoun predictions from models on our task sentences in the absence of any context.
In Section \ref{sec:injecting-an-introductory-context}, we will examine whether models can overcome this bias with reasoning when provided with context establishing a single correct answer.

\begin{cframed}
\noindent\textbf{Example:} \nl{The accountant was asked about \_\_\_ charges for preparing tax returns.}

\noindent \textbf{No single answer} (among \textit{his}, \textit{her}, \textit{their}, \textit{xyr})
\end{cframed}

As we cannot evaluate accuracy on a task with no single correct answer, we show the counts of model predictions of different pronoun declensions in Figure \ref{fig:context-free-model-predictions}, averaged over all models. Model-specific counts are shown in Appendix \ref{sec:context-free-model-predictions-appendix}.
Even though our task sentences are designed such that any pronoun set can be used grammatically, all models tend to assign higher probability to \textit{he/him/his} than other pronoun sets.

Obtaining pronoun predictions without context is a popular method to measure model bias, with numerous papers~\citep[\textit{inter alia}]{kurita-etal-2019-measuring} showing that associations between occupations and pronouns are based on social gender stereotypes, e.g., \emph{doctor}-\emph{he} and \emph{nurse}-\emph{she}.
However, model pronoun predictions might reflect dataset artifacts such as the choice of occupations, or be a statistical accident of the chosen templates~\citep{quantifyingsocialbiases2022}.
In addition, intrinsic biases may not correlate with actual pronoun use with context~\citep{goldfarb-tarrant-etal-2021-intrinsic}.
In order to test for such extrinsic behaviours, the rest of this paper examines whether models can override their intrinsic statistical biases on these same templates when provided with the right pronoun to use.

\begin{figure*}[ht]
    \centering
    \includegraphics[width=\linewidth]{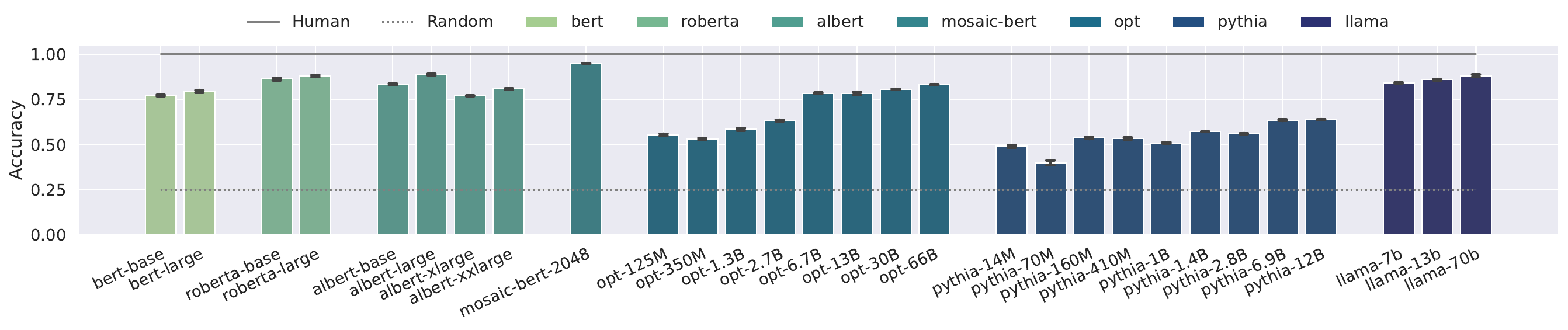}
    \vspace{-8mm}
    \caption{Pronoun fidelity by model with an introductory context. Accuracy is averaged across occupations, pronouns and grammatical cases, and is above chance (0.25) but below human performance (1.0).}
    \label{fig:intro-barplots}
    \vspace{-2mm}
\end{figure*}

\begin{figure*}[ht]
  \begin{tabular}{p{0.16\linewidth}p{0.8\linewidth}}
        \vspace{0.01mm} \small{he/him/his} & \includegraphics[valign=m,width=\linewidth]{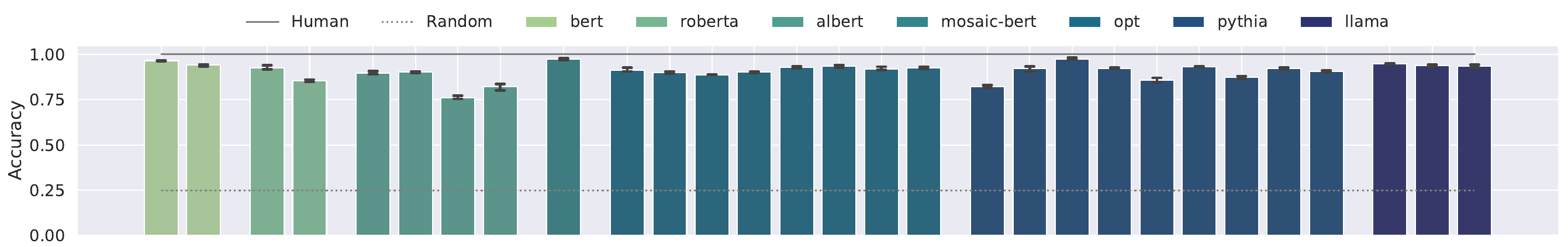} \\
        \vspace{-4mm} \small{she/her/her} \newline \small{(* < he/him/his)} & \includegraphics[valign=m,width=\linewidth]{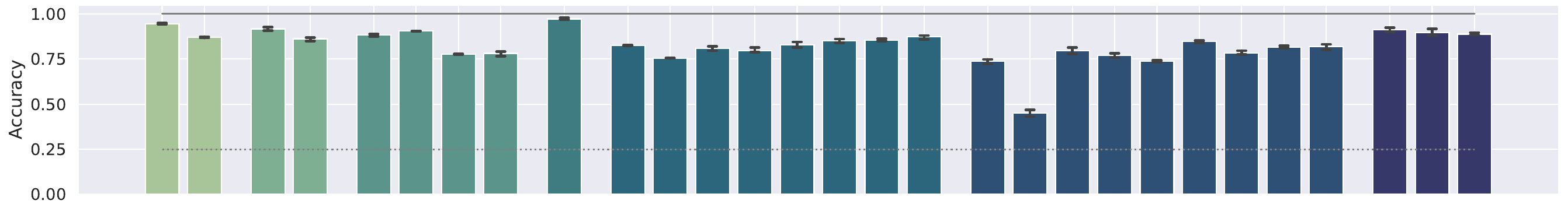} \\
        \vspace{-4mm} \small{they/them/their} \newline \small{(* < she/her/her)} & \includegraphics[valign=m,width=\linewidth]{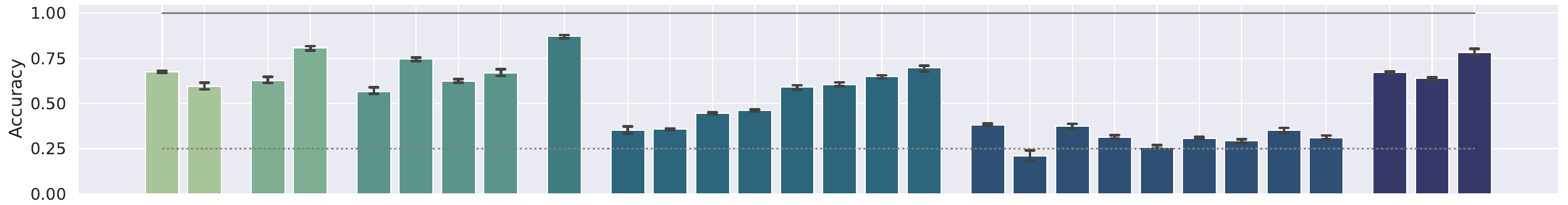} \\
        \vspace{-8mm} \small{xe/xem/xyr} \newline \small{(* < she/her/her)} & \includegraphics[valign=m,width=\linewidth]{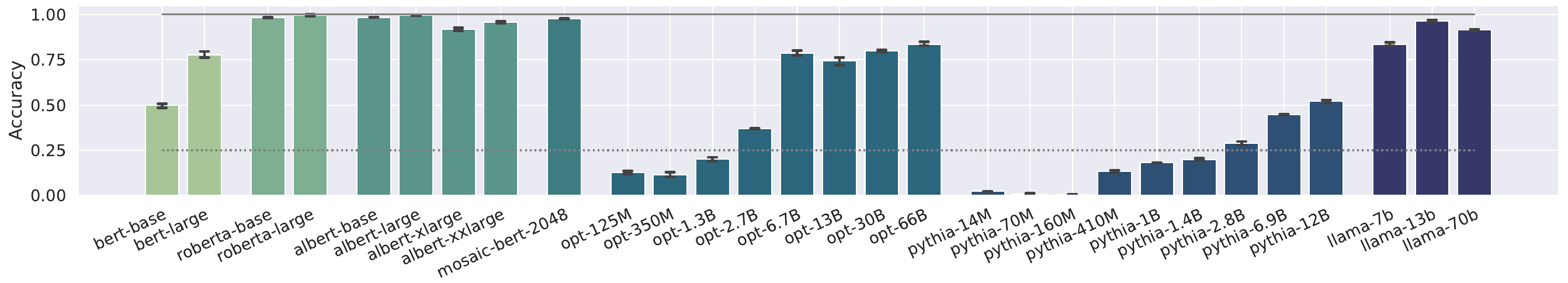} \\
    \end{tabular}
    \caption{Pronoun fidelity by model with an introductory context, split by pronoun series. Model accuracy is compared to chance (0.25) and human performance (1.0) and * denotes statistical significance.}
    \label{fig:intro-pronoun-accuracy}
    \vspace{-4mm}
\end{figure*}

\section{Injecting an Introductory Context}
\label{sec:injecting-an-introductory-context}

When models are provided with an introductory sentence explicitly establishing the pronoun to use for an entity, can they use that pronoun to refer to the same entity in the immediate next sentence?

\begin{cframed}
\noindent\textbf{Example:} \nl{The accountant had just eaten a big meal so \textcolor{forestgreen}{\underline{her}} stomach was full. The accountant was asked about \_\_\_ charges for preparing tax returns.}

\noindent\textbf{Correct answer:} \textcolor{forestgreen}{\underline{her}}
\end{cframed}

As Figure \ref{fig:intro-barplots} shows, \textbf{all models perform better than chance at pronoun fidelity with a simple introduction} (up to 0.95 with \textsc{MosaicBERT}), but not as well as humans, who achieve perfect performance. We also see improvements with increasing model scale, with the exception of \textsc{ALBERT-v2}, as in \citet{tay2022scaling}. \\

\noindent\textbf{Which pronouns are harder?}\quad
Even in the simplest case of the pronoun fidelity task, patterns emerge when split by pronoun, as shown in Figure \ref{fig:intro-pronoun-accuracy}. Overall model \textbf{accuracy on \textit{he/him/his} is significantly higher than \textit{she/her/her}, which in turn is significantly higher than both \textit{they/them/their} and \textit{xe/xem/xyr}}, in line with previous findings that language technology has gaps when it comes to neopronouns~\citep{lauscher-etal-2023-em}.
Models show intriguing patterns with these last two pronoun sets. Most encoder-only models appear to handle the neopronoun better than singular \textit{they} (e.g., \textsc{BERT-large} has an accuracy of 0.78 on \textit{xe/xem/xyr} compared to 0.60 on \textit{they/them/their}), which warrants further investigation.
Decoder-only models smaller than 6.7B parameters struggle with the neopronoun, with every \textsc{OPT} and \textsc{Pythia} model smaller than 2.7B parameters performing below chance, and in some cases (e.g., \textsc{Pythia-14M}, \textsc{Pythia-70M} and \textsc{Pythia-160M}) even performing close to 0.0.
Beyond this scale, however, models perform better on \textit{xe/xem/xyr} than on singular \textit{they}, with \textsc{Llama-13B} achieving 0.96 accuracy on the neopronoun. These differences are statistically significant. As the training data for individual model families is the same, this might suggest that decoder-only models generalize to novel pronouns starting at the scale of 6.7B parameters, but in light of \citet{schaeffer2023emergent}, this result could just as well be a mirage resulting from our use of accuracy, a discontinuous metric.
In either case, our observations could also explain the poor performance that some previous studies of neopronouns find, as the largest model that \citet{hossain-etal-2023-misgendered} experiment with, for instance, is \textsc{OPT-6.7B}.
The lower performance of bigger models with singular \textit{they} could also be a reflection of human processing difficulties with definite, specific singular \textit{they}, as has been observed in linguistics~\citep{Conrod_2019}.

\begin{figure*}[th!]
    \centering
    \begin{subfigure}[b]{0.325\linewidth}
        \centering
        \includegraphics[width=\linewidth]{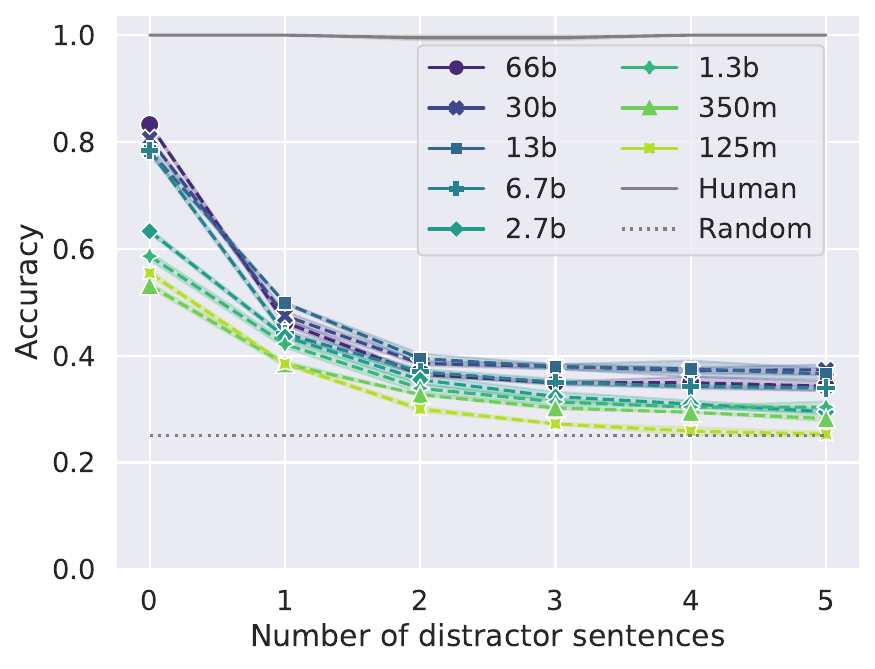}
        \caption{\textsc{OPT}}
    \end{subfigure}
    \hfill
    \begin{subfigure}[b]{0.325\linewidth}
        \centering
        \includegraphics[width=\linewidth]{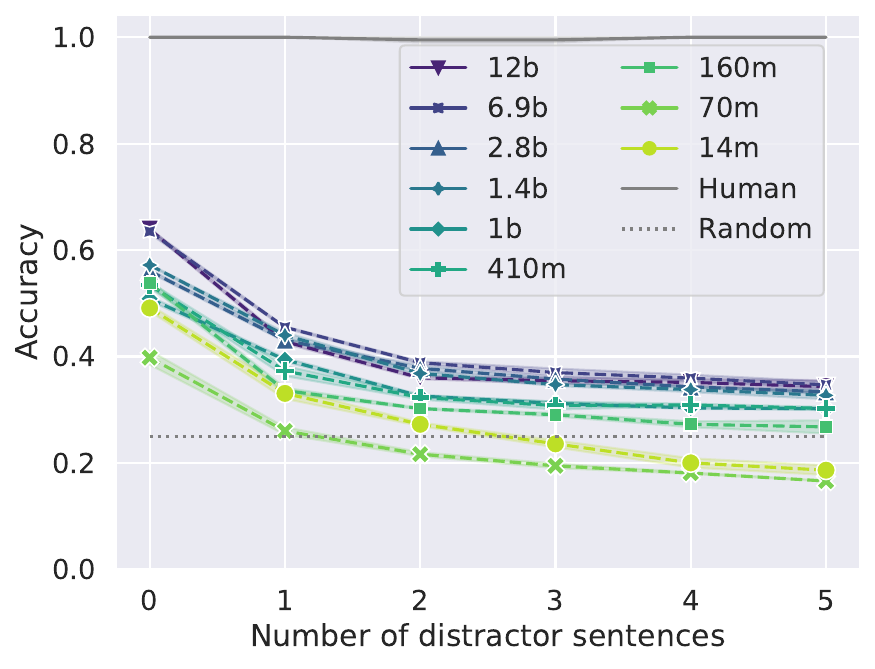}
        \caption{\textsc{Pythia}}
    \end{subfigure}
    \hfill
    \begin{subfigure}[b]{0.325\linewidth}
        \centering
        \includegraphics[width=\linewidth]{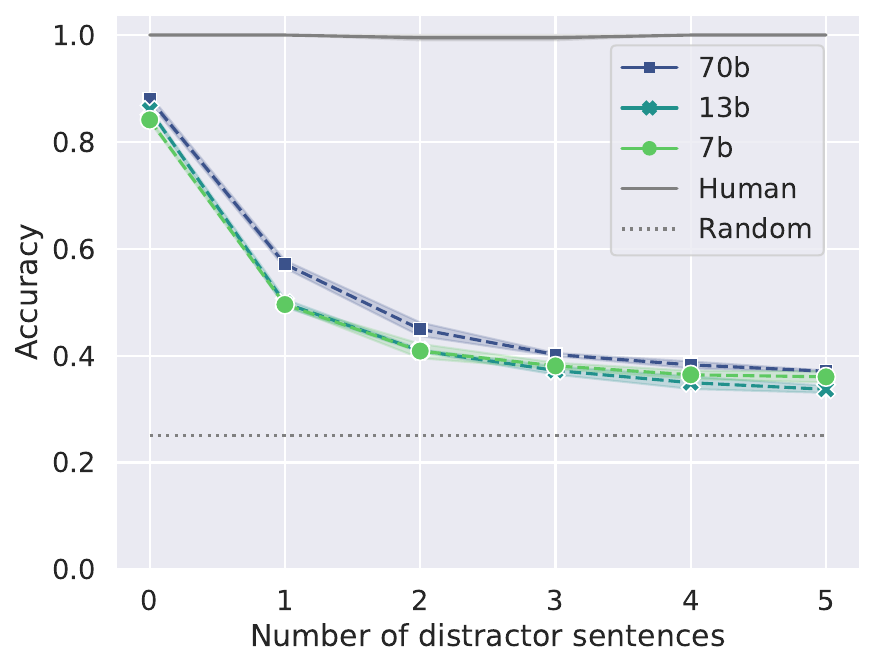}
        \caption{\textsc{Llama-2}}
    \end{subfigure}
    \\
    \begin{subfigure}[b]{0.325\linewidth}
        \centering
        \includegraphics[width=\linewidth]{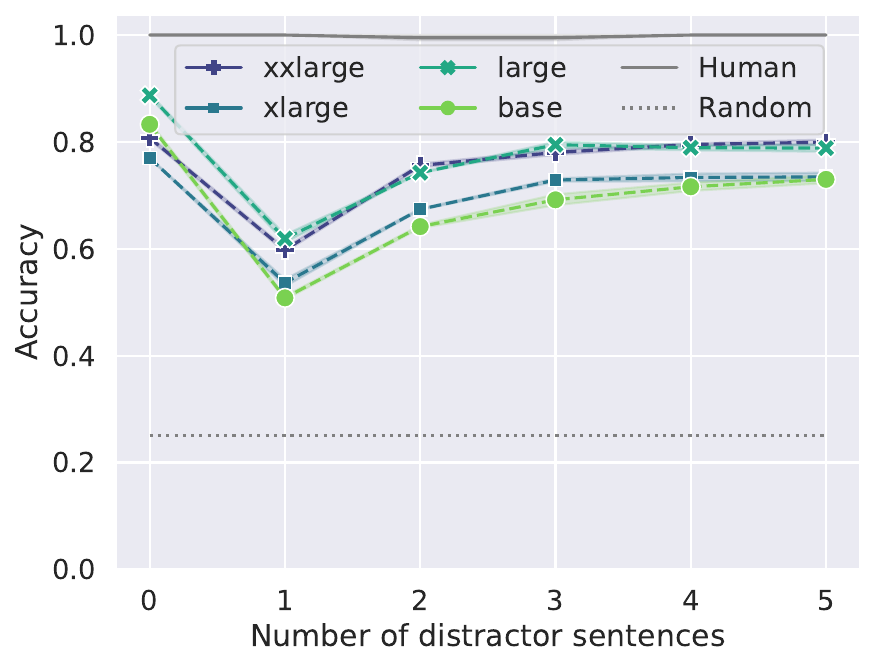}
        \caption{\textsc{ALBERT-v2}}
    \end{subfigure}
    \hfill
    \begin{subfigure}[b]{0.325\linewidth}
        \centering
        \includegraphics[width=\linewidth]{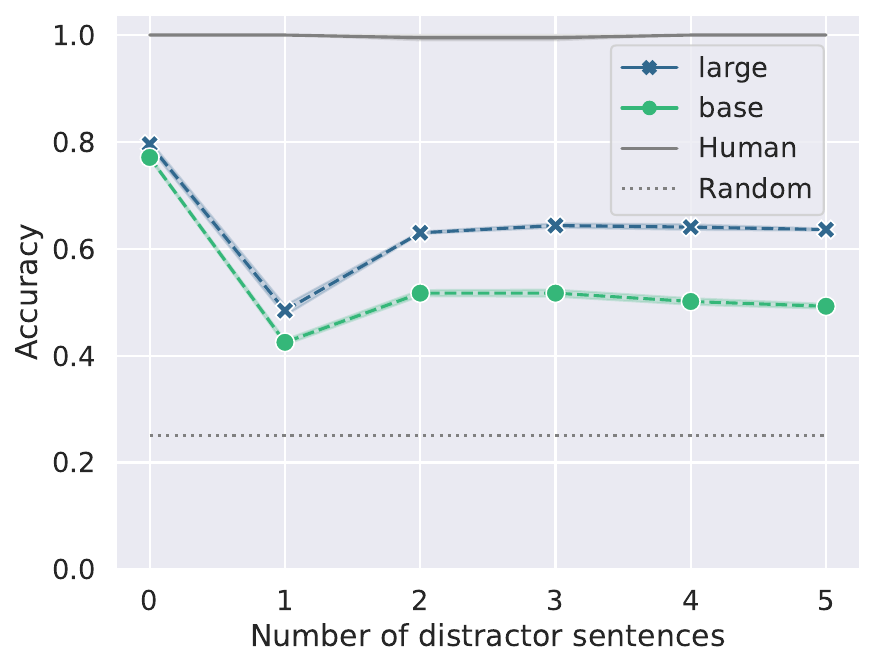}
        \caption{\textsc{BERT}}
    \end{subfigure}
    \hfill
    \begin{subfigure}[b]{0.325\linewidth}
        \centering
        \includegraphics[width=\linewidth]{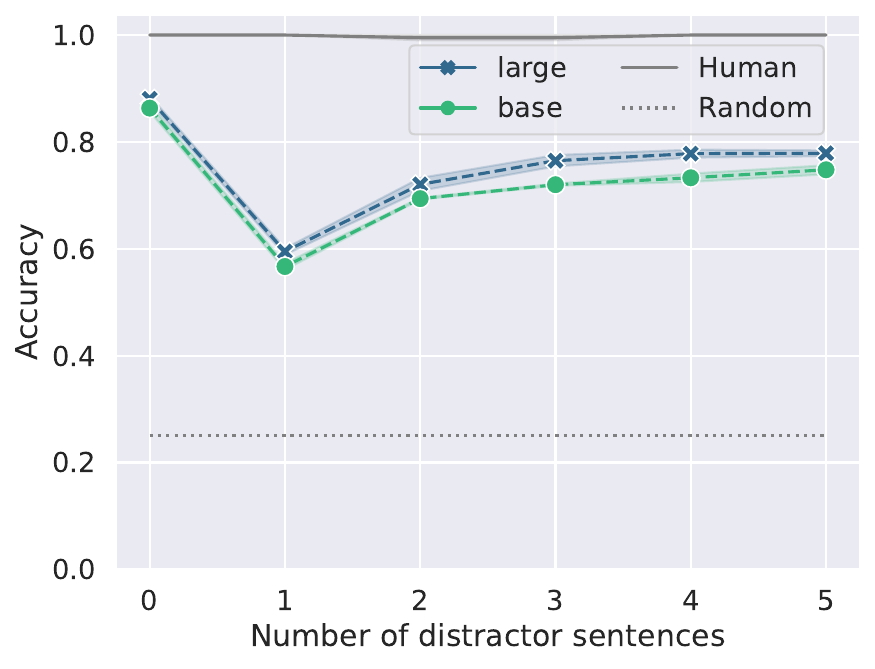}
        \caption{\textsc{RoBERTa}}
    \end{subfigure}
    \caption{With more distractors, decoder-only models (above) get steadily worse; encoder-only models (below) get worse with one distractor and then recover, plateauing below their no-distractor accuracy.}
    \label{fig:scaling-by-model}
    \vspace{-4mm}
\end{figure*}

\begin{figure}[hb!]
    \centering
    \begin{subfigure}[b]{0.91\linewidth}
        \centering
        \includegraphics[width=\textwidth]{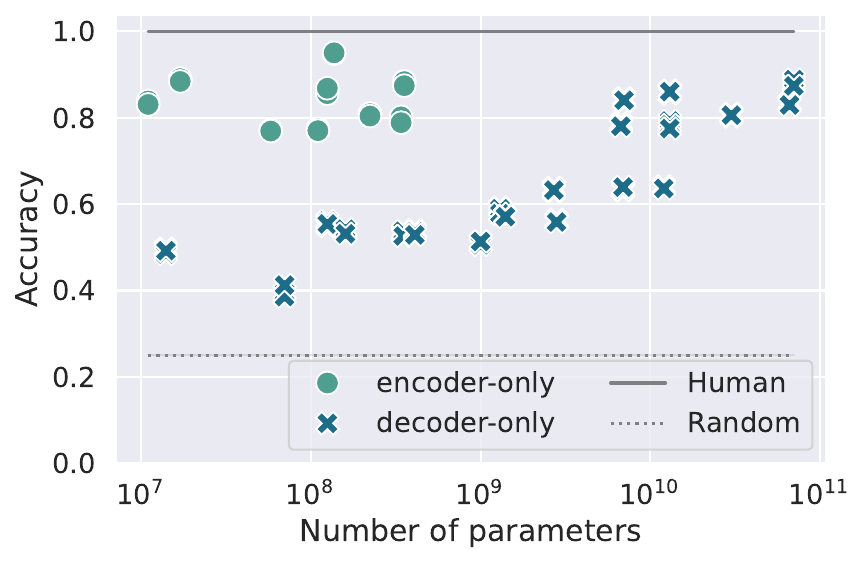}
        \caption{With 0 distractors}
    \end{subfigure}
    \begin{subfigure}[b]{0.91\linewidth}
        \centering
        \includegraphics[width=\textwidth]{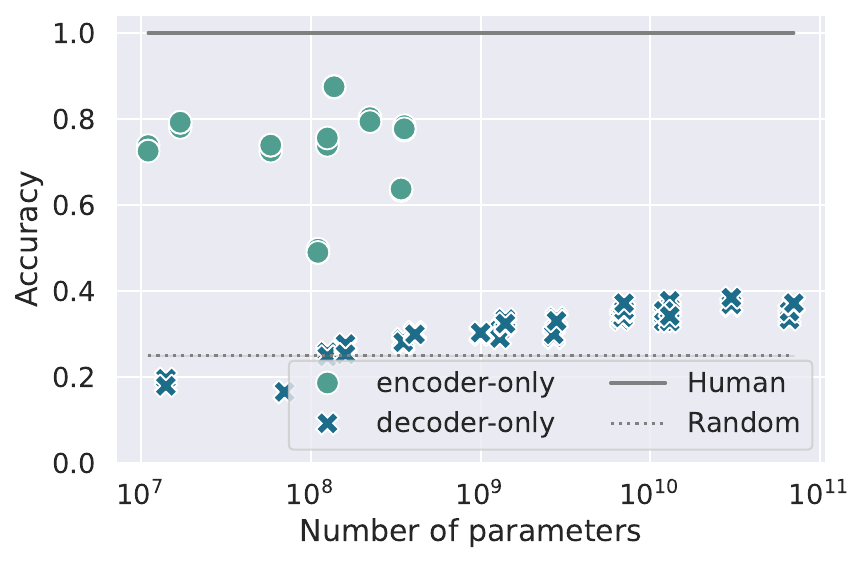}
        \caption{With 5 distractors}
    \end{subfigure}
    \caption{Scaling behaviour by architecture. With 0 distractors (above), encoder-only models are comparable to decoder-only models orders of magnitude larger. With 5 distractors (below), encoder-only models are far better.}
    \label{fig:scaling-by-architecture}
\end{figure}

\section{Adding Distractors}
\label{sec:injecting-distractors}

To further probe whether models actually ``reason'' when provided with context, we systematically inject sentences containing distractor pronouns between the introduction and the task, reflecting a natural usage scenario where multiple people are discussed with definite references and pronouns.

\begin{cframed}

\noindent\textbf{Example:} \nl{The accountant had just eaten a big meal so \textcolor{forestgreen}{\underline{her}} stomach was full. The taxpayer needed coffee because \hl{their} day had started very early. \hl{Their} sleep had been fitful. The accountant was asked about \_\_\_ charges for preparing tax returns.}

\noindent\textbf{Correct answer:} \textcolor{forestgreen}{\underline{her}}

\end{cframed}

Figure \ref{fig:scaling-by-model} shows that distractors degrade performance for all models.
Encoder-only and decoder-only models show different performance curves as more distractors are added:
all decoder-only models get steadily worse, whereas encoder-only models perform the worst with one distractor and then seem to slowly recover, never quite reaching their level of performance with no distractors. Scaling generally holds within model families, with larger models performing better with more distractors than smaller models of the same type. Figure \ref{fig:scaling-by-architecture} examines the interplay of scaling and architecture at a higher level, comparing results on the easiest case of pronoun fidelity (no distractors) with the hardest case (5 distractors). Surprisingly, \textbf{with no distractors, encoder-only models are much better than decoder-only models of the same scale}, and their performance is comparable to or better than decoder-only models that are orders of magnitude larger; \textsc{RoBERTa-base} (125M) is 0.86 accurate compared to \textsc{OPT-125M}'s 0.55, and exceeds \textsc{OPT-66B}'s 0.83 despite being more than 500 times smaller. In the hardest version of our task \textbf{with five distractors, encoder-only models are far better than \textit{all} decoder-only models}, which show dramatically degraded performance; \textsc{Llama-70B} only achieves 0.37 accuracy, compared to \textsc{MosaicBERT}'s impressive 0.87.
The lack of robustness of decoder-only models to distractors is striking, given that most state-of-the-art models today are decoder-only models.
We hypothesize that architectural differences might explain the performance gaps; encoder-only models might use bidirectional attention to more closely relate the entity mentions in the introduction and task sentences.
Training on next token prediction might also make decoder-only models prone to recency bias.\looseness=-1

Using vanilla and chain-of-thought prompting (Appendix \ref{sec:prompting-results}) show the same patterns of degradation, reinforcing that \textbf{model pronoun fidelity is not robust}, and good performance with no distractors (\S\ref{sec:injecting-an-introductory-context}) is likely not due to ``reasoning'' at all.

\begin{figure*}[th!]
    \centering
    \includegraphics[height=0.3cm,keepaspectratio]{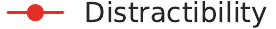}
    \includegraphics[height=0.3cm,keepaspectratio]{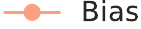} \\
    \begin{subfigure}[b]{0.325\linewidth}
        \centering
        \includegraphics[width=\linewidth]{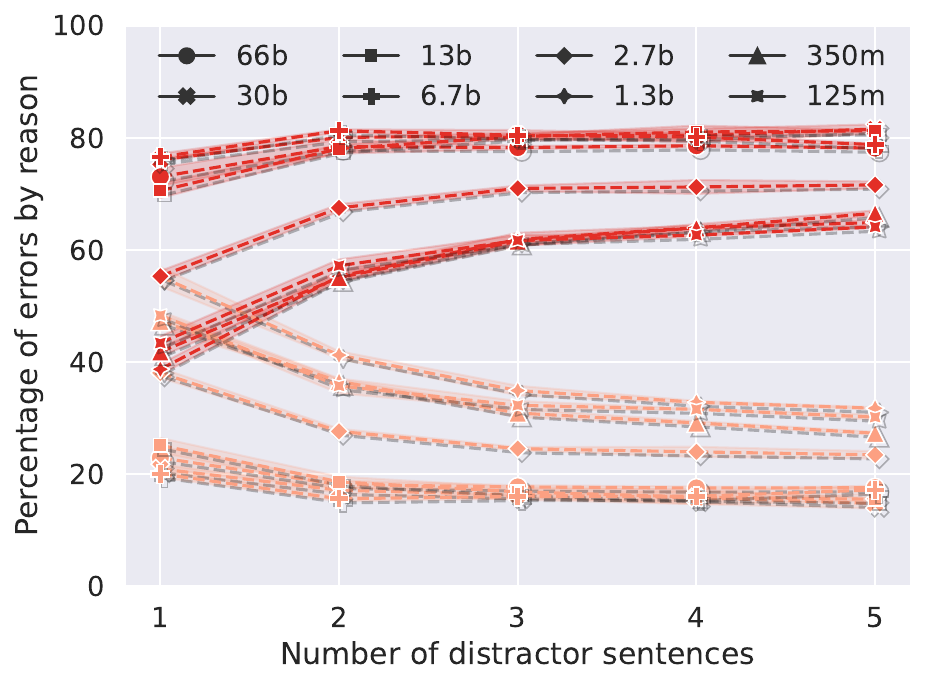}
        \caption{\textsc{OPT}}
    \end{subfigure}
    \hfill
    \begin{subfigure}[b]{0.325\linewidth}
        \centering
        \includegraphics[width=\linewidth]{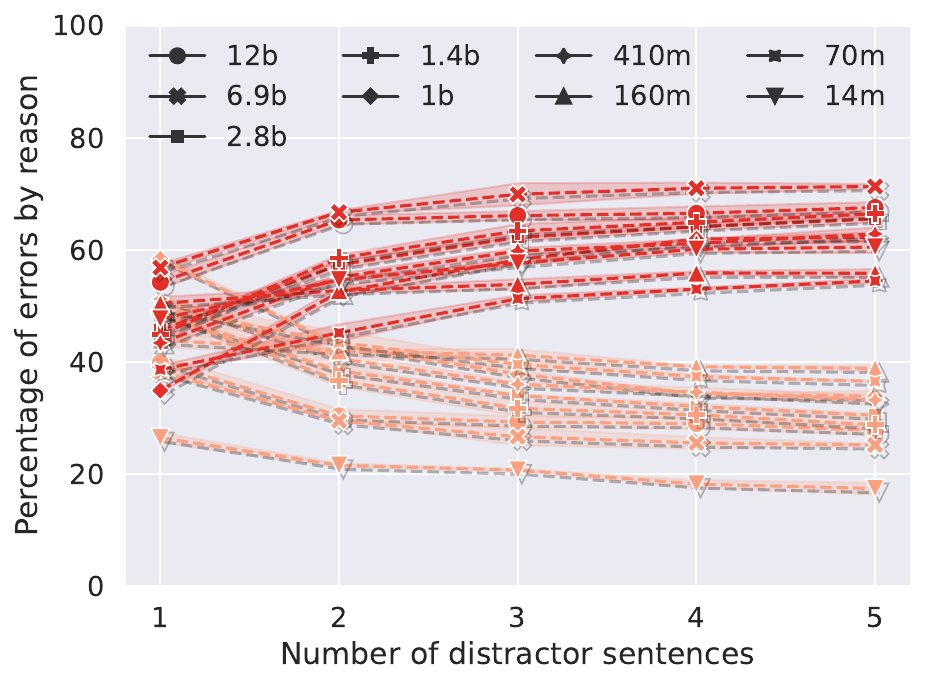}
        \caption{\textsc{Pythia}}
    \end{subfigure}
    \hfill
    \begin{subfigure}[b]{0.325\linewidth}
        \centering
        \includegraphics[width=\linewidth]{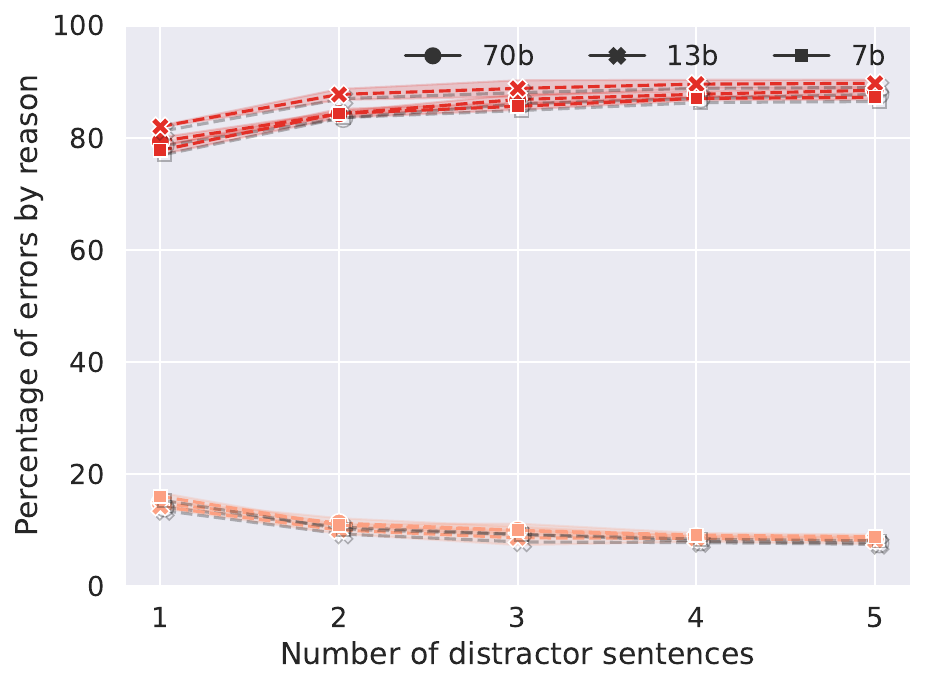}
        \caption{\textsc{Llama-2}}
    \end{subfigure}
    \\
    \begin{subfigure}[b]{0.325\linewidth}
        \centering
        \includegraphics[width=\linewidth]{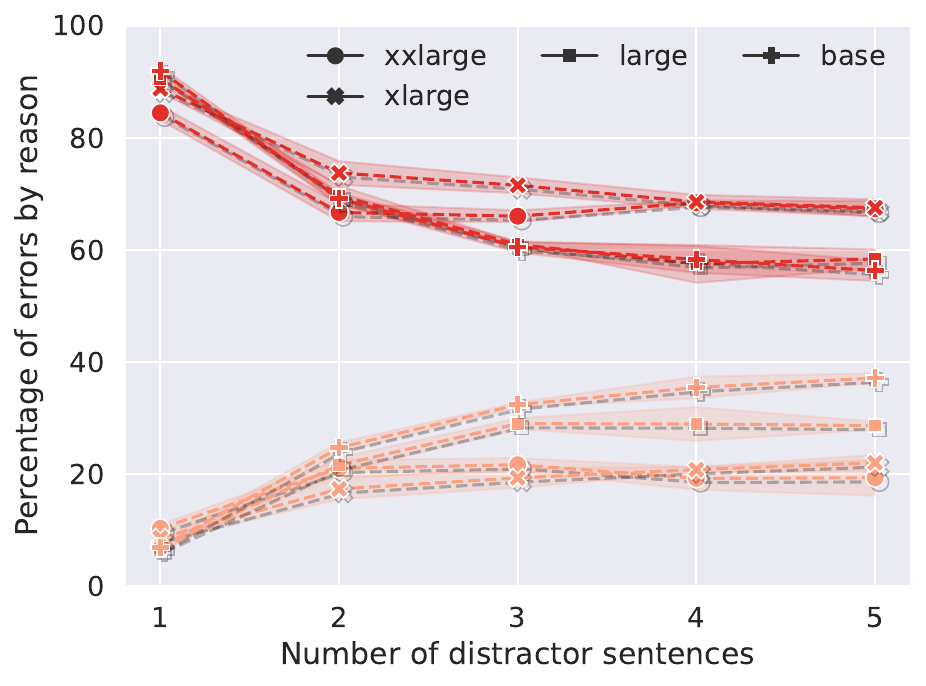}
        \caption{\textsc{ALBERT-v2}}
    \end{subfigure}
    \hfill
    \begin{subfigure}[b]{0.325\linewidth}
        \centering
        \includegraphics[width=\linewidth]{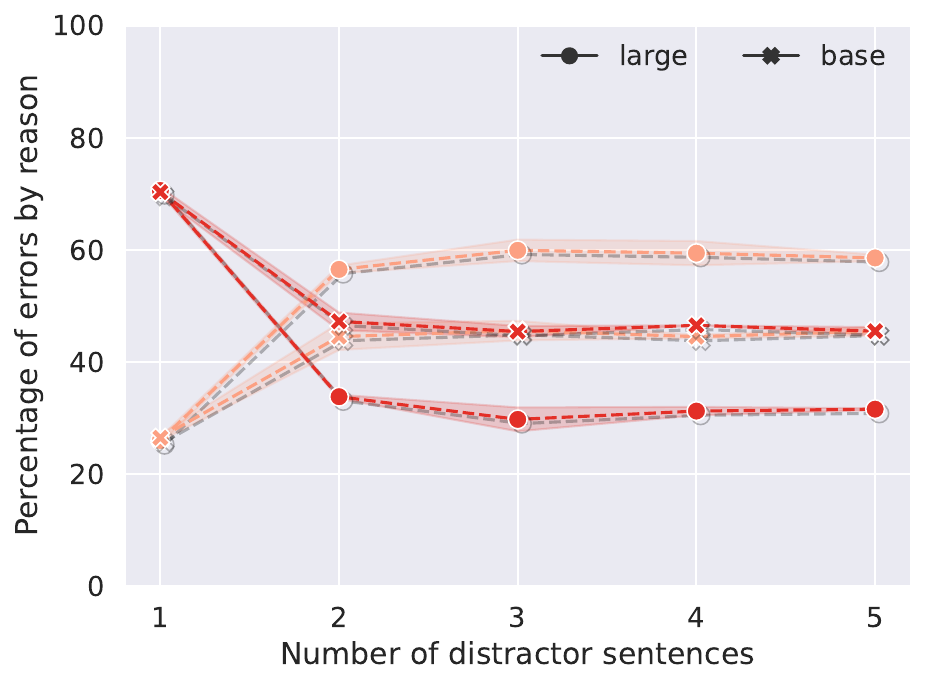}
        \caption{\textsc{BERT}}
    \end{subfigure}
    \hfill
    \begin{subfigure}[b]{0.325\linewidth}
        \centering
        \includegraphics[width=\linewidth]{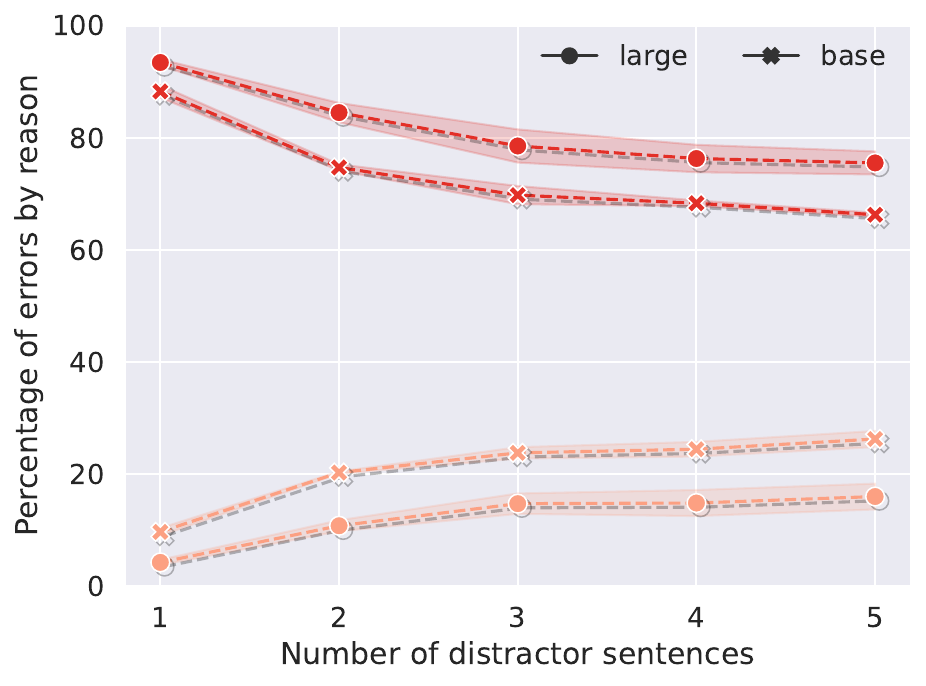}
        \caption{\textsc{RoBERTa}}
    \end{subfigure}
    \caption{Trends in model distractibility (use of the distractor pronoun) and model bias (reverting to the context-free prediction).
    With more distractors, the proportion of errors due to distraction increases for decoder-only models (above) and decreases for encoder-only models (below).}
    \label{fig:distractibility-by-model}
\end{figure*}

\section{Distractibility versus Bias}
\label{error-analysis}

In adding distractor sentences, we add distance from the introduction via additional tokens that might make the model forget the original occupation-pronoun association, and the distractor pronoun also acts as a competing token that the model might accidentally repeat.
In this section, we focus on the error cases to disentangle whether models are ``forgetting'' and reverting to biased predictions from Section \ref{sec:context-free-model-predictions}, or if they are actually being distracted.
When a model gets the answer wrong, it is for one of three reasons: (1) distractibility, i.e., repeating the distractor pronoun, (2) bias, i.e., reverting to the model's context-free prediction, or (3) picking a completely different pronoun. Our example illustrates all three possibilities, and we hypothesize that the first two possibilities are much more frequent than the third.

In cases where the distractor pronoun is the same as the model's context-free prediction, it is impossible to disentangle distractibility and bias just from the model's prediction. Hence, we exclude these and focus on the unambiguous error cases.
As expected, we find that 74-93\% of unambiguous model errors can be attributed to either model distractibility or bias.

\begin{cframed}
\centerline{\textbf{Context-free} (\S\ref{sec:context-free-model-predictions})}

\noindent\textbf{Example:}
\nl{The accountant was asked about \_\_\_ charges for preparing tax returns.}

\noindent\textbf{Prediction:} \textbf{\textcolor{red!85}{his}}

\vspace{0.05in}
\Hrule[gray!40]{3pt}{5pt}
\vspace{0.05in}

\centerline{\textbf{With introduction and distractors} (\S\ref{sec:injecting-distractors})}

\noindent\textbf{Example:} \nl{The accountant had just eaten a big meal so \textcolor{forestgreen}{\underline{her}} stomach was full. The taxpayer needed coffee because \hl{their} day had started very early. \hl{Their} sleep had been fitful. The accountant was asked about \_\_\_ charges for preparing tax returns.}

\noindent\textbf{Correct answer:} \textcolor{forestgreen}{\underline{her}}

\noindent\textbf{Distraction error:} \hl{their}

\noindent\textbf{Bias error:} \textbf{\textcolor{red!85}{his}}

\noindent\textbf{Other error:} xyr

\end{cframed}

We first examine model distractibility, i.e., what percentage of errors are caused by models repeating the distractor pronoun instead of the correct pronoun.
As expected, Figure \ref{fig:distractibility-by-model} shows that across models, distraction is indeed the primary type of error for most models.
\textbf{Decoder-only models get increasingly distracted with more distractors}, i.e., the proportion of errors due to distractor pronoun repetition steadily increases as distractors are added, saturating just below 85\%.
On the other hand, \textbf{encoder-only models seem to become \textit{less} distractible} with the addition of more distractors.
We know from the previous section that encoder-only models recover in their pronoun fidelity with 2-5 distractors, but here we measure distractibility as a percentage of all errors.
Thus, a constant or increasing proportion of all the model errors could be due to distraction, and the fact that it \textit{isn't} for encoder-only models is quite surprising!
We leave it to future work to investigate whether this behaviour relates to positional bias or context use.\looseness=-1

As their proportion of distraction errors goes down, \textbf{encoder-only models increasingly revert to biased predictions}. With \textsc{BERT-large} in particular, as soon as there is more than one distractor, the biggest proportion of errors is due to bias rather than distraction. \textsc{BERT-large} appears more  biased and less distractible than \textsc{BERT-base}, in contrast to all other models. Generally, larger models seem to be more distractible and revert to their bias less often, whereas smaller models are more biased and less distractible.
Our findings on bias errors contrast with \citet{tal-etal-2022-fewer}, where larger models make a higher proportion of bias errors on a downstream task than smaller models. This might be due to our task having distractors, which seem to strongly influence model behaviour in this setting.

The high distractibility of all models shows that \textbf{models are not robust reasoners}, and the contrast in error behaviour between encoder-only and decoder-only models further highlights their differences.
This shows that claims about decoder-only models should not be applied to all LLMs, and that reasoning must be evaluated carefully, accounting for the possibility of inflated performance due to shallow heuristics like repetition.

\section{Discussion and Future Work}

Our results show that even the biggest models of today are not up to the task of pronoun fidelity once it includes a single sentence discussing another person.
All models are easily distracted, but encoder-only models and decoder-only models show very different patterns both in performance degradation with more distractors and their reasons for errors.
Performance on this type of reasoning task should be evaluated carefully, with attention to how the overall patterns break down by different pronouns, and accounting for the possibility of repetition.
Below we expand on some questions raised by our findings. \\

\noindent \textbf{Improving robust pronoun fidelity.}\quad
A natural direction of future work is to solve the problem of robust pronoun fidelity, particularly in decoder-only models, which are unlikely to be replaced by encoder-only models with poorer generation abilities.
A promising direction might be to encourage models to explicitly track associations between individuals and pronoun sets, just as people do.
In fact, prior work has noted success with generative models when explicitly tracking mentions of entities across multiple tasks~\citep{ji-etal-2017-dynamic} and in the context of story generation~\citep{fan-etal-2019-strategies}.
We urge researchers interested in this direction to treat \textsc{RUFF} as an evaluation dataset, as it was designed.
Due to the presence of positional and associative heuristics (see Appendix \ref{sec:limitations}), \textsc{RUFF} should not be seen as a source of data for fine-tuning or in-context learning, which is also why we do not run these experiments. \\

\noindent \textbf{On ``reasoning.''}\quad
Throughout the paper, we refer to ``reasoning,'' but this is inaccurate.
Even the higher performance of encoder-only models cannot accurately be attributed to ``reasoning'' in the same way that we use this word for humans, as these models are not grounded in \textit{meaning} from the real world~\citep{bender-koller-2020-climbing}.
We use the word reasoning in line with other work in the field, but note that as these are all language models, it is more accurate to say that the way that decoder-only models model language is prone to repetition---or stochastic parroting~\citep{stochastic_parrots}---of recent examples of the same word class, compared to encoder-only models. \\

\noindent \textbf{\textit{Why} exactly do we see the patterns we see?}\quad
Our dataset design and error analysis shed light on model \textit{behaviour}, allowing us to evaluate different architectures comparably and disentangle the effects of repetition, distraction and statistical bias.
However, it is beyond the scope of this paper to investigate where in the model architecture, neurons or pre-training data this comes from and what we can do about it towards improving reasoning and mitigating bias. Tools for model interpretability, e.g., attribution analysis, could help here, and are an important direction for future work. \\

\noindent \textbf{Beyond our dataset.}\quad
Given the breadth of our task definition, future work could examine pronoun fidelity in other contexts, e.g., for participants, for names by extending \citet{hossain-etal-2023-misgendered}, with differently ordered sentences, with real-world data as in \citet{webster-etal-2018-mind} and \citet{levy-etal-2021-collecting-large}, and in domains beyond simple narratives~\citep{pradhan-etal-2013-towards}.
Additionally, we evaluate on a version of this task that allows us to quantify repetition, i.e., the grammatical case of the elicited pronoun is the same as the case shown in the context. Examining model performance where a pronoun is shown in one grammatical case and then elicited in a different one would be interesting to probe syntactic generalization.

\section{Related Work}

\noindent \textbf{Pronoun fidelity.}\quad
\citet{hossain-etal-2023-misgendered} and \citet{ovalle23TANGO} both study pronoun fidelity when models are prompted with a pronoun series to use for an individual, but they only consider simplistic pronoun use with no more than one person at a time.
Although we look at within-language pronoun use, faithful pronoun use in context has also been studied in machine translation~\citep{muller-etal-2018-large,voita-etal-2018-context,fernandes-etal-2023-translation}, where there is also a ground truth. Similar to our work, \citet{sharma-etal-2022-sensitive} injects context with an explicit coreference to encourage faithful pronoun translation. 
However, none of these papers explore the \textit{robustness} of pronoun fidelity in the presence of distractors. \\

\noindent \textbf{Reasoning with pronouns.}\quad
Most existing work about LLM reasoning with pronouns focuses on the task of coreference resolution, i.e., the ability to \textit{identify} the connection between a pronoun and an entity, which may not translate to \textit{faithful reuse} of that pronoun later, as in our work.
Reasoning with pronouns typically uses Winograd schemas~\citep{levesque2012,abdou-etal-2020-sensitivity,emelin-sennrich-2021-wino},
or Winograd-like schemas about named individuals \citep{webster-etal-2018-mind,zhao-etal-2018-gender}, or people referred to by their occupation~\citep{rudinger-etal-2018-gender,levy-etal-2021-collecting-large}. Most studies focus on \textit{he} and \textit{she}, but recent work has expanded to include singular \textit{they}~\citep{baumler-rudinger-2022-recognition} and neopronouns~\citep{cao-daume-iii-2021-toward,felkner-etal-2023-winoqueer}, as we do. \\

\noindent \textbf{Pronouns and occupational bias.}\quad
Stereotypical associations between pronouns and occupations have been studied in masked token prediction~\citep{kurita-etal-2019-measuring,de-vassimon-manela-etal-2021-stereotype,tal-etal-2022-fewer} and embeddings~\citep{bolukbasiwordemb2016,zhao-etal-2019-gender}, but these studies typically use brittle methodology~\citep{gonen-goldberg-2019-lipstick,quantifyingsocialbiases2022} and measure intrinsic bias, which may not translate to extrinsic bias or harms~\citep{goldfarb-tarrant-etal-2021-intrinsic}.
Unlike these works, we evaluate extrinsic bias and performance through our focus on natural pronoun use in context. \\

\noindent \textbf{Robustness in context.}\quad
The impact of context on the robustness of language model reasoning has been investigated in many areas other than pronoun fidelity, e.g., negation~\citep{gubelmann-handschuh-2022-context}, linguistic acceptability~\citep{sinha-etal-2023-language}, natural language inference~\citep{srikanth-rudinger-2022-partial}, and question answering ~\citep{10.1162/tacl_a_00638,levy2024task}.

\section{Conclusion}
We introduce the task of pronoun fidelity to evaluate robust, faithful and harm-free pronoun use in language models, and we present \textsc{RUFF}, a dataset we designed to evaluate it.
We find evidence of faithful pronoun use only in a very simple setting, i.e., when only one person is discussed.
Even here, models show significant performance disparities with neopronouns, singular \textit{they} and \textit{she/her/her}, compared to \textit{he/him/his}.
Even adding a single sentence about a second individual with a different pronoun causes accuracy to drop dramatically, showing that pronoun fidelity is neither robust to non-adversarial distractors nor due to ``reasoning.''
As more distractor sentences are added, encoder-only models perform better overall, but increasingly revert to biased predictions, while decoder-only models get increasingly distracted.
Our results show that in a setting that is very simple for humans, widely-used large language models are unable to robustly and faithfully reason about pronouns, and continue to amplify discrimination against users of certain pronouns.
We encourage researchers to bridge the performance gaps we report and to more carefully evaluate ``reasoning,'' especially when simple repetition could inflate perceptions of model performance.

\section{Limitations}
\label{sec:limitations}

\noindent \textbf{Shallow heuristics.}\quad
Much of the recent progress on reasoning datasets has been critically investigated and shown to often be a result of spurious correlations and dataset artifacts~\citep{trichelair-etal-2019-reasonable,elazar-etal-2021-back}. We caution readers that our dataset also gives a very \textit{generous} estimate of model reasoning performance, as many of our task sentences are not ``Google-proof'' \citep{levesque2012}, i.e., they can be solved with shallow heuristics such as word co-occurrences. Consider the following task sentence: \textit{The janitor said not to step on the wet floor, otherwise \_\_\_ would have to mop it all over again}. \textit{Janitor} is more strongly associated with \textit{mop} than \textit{child}, which could easily be exploited by models to solve the dataset without solving the task with something resembling ``reasoning.''
Another shallow heuristic that can be used to solve our current dataset is to simply return the first pronoun in the context, which happens to always be the correct answer. Our dataset design is flexible and allows for the creation of other orderings of sentences, but this is another example of why our dataset in its current form should only be used as an evaluation dataset, and models should not be pre-trained or fine-tuned with any splits of our data, nor provided with examples for in-context learning. \\

\noindent \textbf{Whose bias?}\quad
Our task as it is defined in Section \ref{sec:task-definition} is much broader than the scope of our dataset. We focus on occupations due to the wide attention they have received in prior literature, but we continue a long tradition of ignoring biases relating to the participants, e.g., \textit{child}, \textit{taxpayer}, etc.
In addition, pronoun fidelity is only one dimension of inclusive language model behaviour, and indeed only one way in which misgendering occurs in language, even in morphologically poor languages such as English. \\

\noindent \textbf{Data contamination.}\quad
We take steps to prevent data contamination following \citet{jacovi-etal-2023-stop}, including not releasing our data in plain text, and not evaluating with models behind closed APIs that do not guarantee that our data will not be used to train future models. However, as we cannot guarantee a complete absence of data leakage unless we never release the dataset, we encourage caution in interpreting results on \textsc{RUFF} with models trained on data after March 2024.

\section*{Acknowledgements}
The authors thank Timm Dill for several rounds of patient annotation, as well as Aaron Mueller, Marius Mosbach, Vlad Niculae, Yanai Elazar, our action editor Hai Zhao, and our anonymous TACL reviewers, for feedback on math, plots and framing, that improved this work.
Vagrant Gautam received funding from the BMBF’s (German Federal Ministry of Education and Research) SLIK project under the grant 01IS22015C. Anne Lauscher's work is funded under the Excellence Strategy of the German Federal Government and States.

\bibliography{custom}
\bibliographystyle{acl_natbib}

\onecolumn

\appendix

\section{List of Occupations}
\label{sec:occupation-list}

The occupations along with their respective participants in parentheses are listed below in alphabetical order. This list is identical to the occupations and participants in \citet{rudinger-etal-2018-gender}, except that we pair examiner with intern rather than victim: \\

\noindent accountant (taxpayer), administrator (undergraduate), advisor (advisee), appraiser (buyer), architect (student), auditor (taxpayer), baker (customer), bartender (customer), broker (client), carpenter (onlooker), cashier (customer), chef (guest), chemist (visitor), clerk (customer), counselor (patient), dietitian (client), dispatcher (bystander), doctor (patient), educator (student), electrician (homeowner), engineer (client), examiner (intern), firefighter (child), hairdresser (client), hygienist (patient), inspector (homeowner), instructor (student), investigator (witness), janitor (child), lawyer (witness), librarian (child), machinist (child), manager (customer), mechanic (customer) nurse (patient), nutritionist (patient), officer (protester), painter (customer), paralegal (client), paramedic (passenger), pathologist (victim), pharmacist (patient), physician (patient), planner (resident), plumber (homeowner), practitioner (patient), programmer (student), psychologist (patient), receptionist (visitor), salesperson (customer), scientist (undergraduate), secretary (visitor), specialist (patient), supervisor (employee), surgeon (child), teacher (student), technician (customer), therapist (teenager), veterinarian (owner), worker (pedestrian) 

\section{Context Template Construction}
\label{sec:context-template-details}

For each grammatical case, we create 10 explicit templates, which explicitly demonstrate the coreference between an individual and a pronoun using a subordinate clause, and 10 implicit templates, simple sentences which only contain a pronoun as the subject.
An introduction and the first distractor are always sampled from the explicit templates, and the rest are sampled from the implicit templates, as this reflects natural and coherent use of pronouns in discourse.

For both the explicit and implicit cases, we create five templates with terms with positive connotations (e.g., full, happy) and five templates with the opposite polarity (i.e., hungry, unhappy). We denote $exp\_pos_i$ as the i-th positive explicit template where i ranges from 1 to 5; $exp\_neg_i$ is the corresponding negative version. The introduction template can be selected from any of these 10 possibilities and filled with one of four pronouns.

After this, we pick a first distractor template, limiting ourselves to the five templates of the opposite sentiment of what we first picked, and also excluding the template of the same index and opposite polarity. For example, if we chose $exp\_pos_3$ as our introductory template, we would choose our first distractor template from $\{exp\_neg_1, exp\_neg_2, exp\_neg_4, exp\_neg_5\}$.

After making a choice for the first distractor template, we fill it with any of the three remaining pronouns and then we remove this template's index from our pool, but re-add the index of the introductory template. This is because subsequent distractor templates always use implicit templates. For example, if we chose $exp\_neg_4$ as our first distractor template, we would now choose from $\{imp\_neg_1, imp\_neg_2, imp\_neg_3, imp\_neg_5\}$. For subsequent distractor templates, we sample without replacement from these implicit templates.

\section{Annotator Demographics}
\label{sec:annotator-information}

All three annotators (two authors and an additional annotator) are fluent English speakers.
The two authors who create and validate templates have linguistic training at the undergraduate level.
One author and one annotator have experience with using singular \textit{they} and neopronouns, while the other author has prior exposure to singular \textit{they} but not the neopronoun \textit{xe}.

\section{Annotation Instructions}
\label{sec:annotation-instructions}

\subsection{Task 1 Description}

Together with this annotation protocol, you have received a link to a Google Sheet. The sheet contains 2 data columns and 2 task columns of randomized data. The data columns consist of
\begin{itemize}
    \item Sentences which you are asked to annotate for grammaticality; and
    \item Questions about pronouns in the sentence, which you are asked to answer
\end{itemize}
Please be precise in your assignments and do not reorder the data. The columns have built-in data validation and we will perform further tests to check for consistent annotation.

\subsubsection{Grammaticality}
In the “Grammatical?” column, please enter your grammaticality judgments of the sentence, according to Standard English. The annotation options are:
\begin{itemize}
    \item \textbf{grammatical} (for fluent, syntactically valid and semantically plausible sentences)
    \item \textbf{ungrammatical} (for sentences that have any typos, grammatical issues, or if the sentence describes a situation that don’t make sense, or just sounds weird)
    \item \textbf{not sure} (if you are not sure whether it is clearly grammatical or ungrammatical)
\end{itemize}
Examples:
\begin{itemize}
    \item \textit{The driver told the passenger that he could pay for the ride with cash.}\\
    => grammatical
    \item \textit{The driver said the passenger that he could pay for the ride with cash.}\\
    => ungrammatical (because ‘said’ is intransitive in Standard English)
\end{itemize}

\subsubsection{Questions about pronouns}
Every sentence contains a pronoun, and the “Question” column asks whether it refers to a person mentioned in the sentence or not. The annotation options are:
\begin{itemize}
    \item \textbf{yes} (if the pronoun refers to the person)
    \item \textbf{no} (if the pronoun does not refer to the person)
    \item \textbf{not sure} (if you are not sure about whether the pronoun refers to the person)
\end{itemize}
Examples: 
\begin{itemize}
    \item \textit{The driver told the passenger that he could pay for the ride with cash.}\\
    Does the pronoun he refer to the driver?\\
    => no
    \item \textit{The driver told the passenger that he could pay for the ride with cash.}\\
    Does the pronoun he refer to the passenger?\\
    => yes
\end{itemize}

\subsection{Task 2 Description}
Together with this annotation protocol, you have received a link to a Google Sheet. The sheet contains 1 randomized data column and 1 task column.\\
Each row in the data column consists of multiple sentences, of which precisely one sentence contains a blank. Your task is to determine the appropriate pronoun to fill in the blank, and enter it in the “Pronoun” column. Here, appropriate means correct in both form and case.\\
The tasks are designed to be unambiguous, so please provide only one solution and do not reorder the data.

Example: 
\begin{itemize}
    \item \textit{The driver felt unhappy because he did not make enough money. The driver wondered whether} {\_}{\_}{\_} \textit{should take out a loan.}\\
    => he
\end{itemize}

\section{Experimental Details}
\label{sec:experimental-details}

We use one 40GB NVIDIA A100 GPU for inference with most models, but we require two GPUs for \textsc{OPT-30B} and four for \textsc{OPT-66B} and the \textsc{Llama-2-70B} base and chat models. We access all models using the Huggingface Transformers library \citep{wolf-etal-2020-transformers} and use the minicons library for pseudo log likelihood evaluation \citep{misra2022minicons}.

\section{Prompting}
\label{sec:prompting}

Table~\ref{tab:prompting-templates} shows all 10 prompt templates we use to present our task instances to chat models. The first five are open-ended and the last five provide the model with pronoun options to choose from. These templates are used within vanilla and chain-of-thought prompting paradigms as described below.

\begin{table*}[ht!]
\centering
    \begin{tabular}{ll}
    \toprule
    \textbf{ID}  & \textbf{Template}  \\ 
    \midrule
    0      
             &  \texttt{\{task\}}\slashn\slashn What pronoun should be used to fill the blank? \\
    \midrule
    1      
             &  \texttt{\{task\}}\slashn\slashn The best pronoun to fill in the blank is \\
    \midrule
    2      
             &  Fill in the blank with the correct pronoun.\slashn\slashn\texttt{\{task\}} \\
    \midrule
    3      
             &  Fill in the blank with the appropriate pronoun.\slashn\slashn\texttt{\{task\}} \\
    \midrule
    4      
             &  What pronoun should be used to fill the blank?\slashn\slashn\texttt{\{task\}} \\
    \midrule
    5      
             &  \texttt{\{task\}}\slashn\slashn What pronoun should be used to fill the blank?\slashn\texttt{\{options\}} \\
    \midrule
    6      
             &  \texttt{\{task\}}\slashn\slashn The best pronoun to fill in the blank is\slashn\texttt{\{options\}} \\
    \midrule
    7      
             &  Fill in the blank with the correct pronoun.\slashn\slashn\texttt{\{task\}}\slashn\texttt{\{options\}} \\
    \midrule
    8      
             &  Fill in the blank with the appropriate pronoun.\slashn\slashn\texttt{\{task\}}\slashn\texttt{\{options\}}\\
    \midrule
    9      
             &  What pronoun should be used to fill the blank?\slashn\slashn\texttt{\{task\}}\slashn\texttt{\{options\}} \\
    \bottomrule
    \end{tabular}
\caption{Prompting templates for chat models, where ``task'' is filled with each dataset instance, and ``options'' is a list of four pronouns to choose from, all in the correct case.}
\label{tab:prompting-templates}
\end{table*}

\begin{table*}[ht]
\centering
    \begin{tabular}{ll}
    \toprule
    \textbf{Model type}  & \textbf{Example input}  \\ 
    \midrule
    \textsc{FLAN-T5}      
             &  Fill in the blank with the appropriate pronoun. \\

& \\
& The accountant had just eaten
a big meal so her stomach \\
& was full. The accountant was asked
about \_\_\_ charges for \\
& preparing tax returns. \\
& OPTIONS: \\
& - his \\
& - her \\
& - their \\
& - xyr \\
    \midrule
    \textsc{Llama-2-chat}      
             &  [INST] Fill in the blank with the appropriate pronoun. \\

& \\
& The accountant had just eaten
a big meal so her stomach \\
& was full. The accountant was asked
about \_\_\_ charges for \\
& preparing tax returns. \\
& OPTIONS: \\
& - his \\
& - her \\
& - their \\
& - xyr  [/INST]\\
    \bottomrule
    \end{tabular}
\caption{Example input using template 8 for \textsc{FLAN-T5} and \textsc{Llama-2-chat} models.}
\label{tab:prompting-input-examples}
\end{table*}

\begin{table*}[ht]
\centering
    \begin{tabular}{ll}
    \toprule
    \textbf{Chain-of-thought prompt}  & \textbf{Example input}  \\ 
    \midrule
    \citet{Kojima2022stepbystep}
    &  Q: Fill in the blank with the appropriate pronoun. \\
(eliciting chain of thought)
& \\
& The accountant had just eaten
a big meal so her stomach \\
& was full. The accountant was asked
about \_\_\_ charges for \\
& preparing tax returns. \\
& A: Let's think step by step. \\
    \midrule
    \citet{Kojima2022stepbystep}
    &  Q: Fill in the blank with the appropriate pronoun. \\
(eliciting final answer)
& \\
& The accountant had just eaten
a big meal so her stomach \\
& was full. The accountant was asked
about \_\_\_ charges for \\
& preparing tax returns. \\
& A: Let's think step by step. \texttt{\{generated chain of thought\}} \\
& Therefore, the correct pronoun is \\
\midrule
    \citet{zhou2023large}
    &  Q: Fill in the blank with the appropriate pronoun. \\
(eliciting chain of thought)
& \\
& The accountant had just eaten
a big meal so her stomach \\
& was full. The accountant was asked
about \_\_\_ charges for \\
& preparing tax returns. \\
& A: Let’s
work this out in a step by step way to be sure we \\
& have the right answer. \\
    \midrule
    \citet{zhou2023large}
    &  Q: Fill in the blank with the appropriate pronoun. \\
(eliciting final answer)
& \\
& The accountant had just eaten
a big meal so her stomach \\
& was full. The accountant was asked
about \_\_\_ charges for \\
& preparing tax returns. \\
& A: Let’s
work this out in a step by step way to be sure we \\
& have the right answer. \texttt{\{generated chain of thought\}} \\
& Therefore, the correct pronoun is \\
    \bottomrule
    \end{tabular}
\caption{Example input using template 3 for evaluating \textsc{FLAN-T5-xxl} with two types of chain-of-thought prompting. Prompting happens in two phases regardless of the choice of prompt: eliciting the chain of thought and eliciting the final answer.}
\label{tab:cot-prompting-input-examples}
\end{table*}

\subsection{Vanilla prompting}

With \textsc{FLAN-T5}, vanilla prompting only requires instantiating templates with task instances, whereas \textsc{Llama-2-chat} requires special formatting with \texttt{INST}.
Instantiated examples of one template are shown for both models in Table~\ref{tab:prompting-input-examples}.
The number of maximum new tokens is set to 5 for \textsc{FLAN-T5} and 20 for \textsc{Llama-2-chat} based on experimentation.

\subsection{Chain-of-thought prompting}

We focus on \textsc{FLAN-T5-xxl} for chain-of-thought experiments, and use the strong zero-shot prompts for reasoning proposed by \citet{Kojima2022stepbystep} (\textit{``Let's think step by step''}) and \citet{zhou2023large} (\textit{``Let's work this out in a step by step way to be sure we have the right answer''}), which we append after the template.
Following their codebases, we first allow the models to generate a chain of thought (with 128 maximum new tokens).
Then, we append the chain of thought after the question and elicit the final answer with the string \textit{``Therefore, the correct pronoun is,''} allowing the model to generate up to 10 new tokens.
This two-step process is illustrated with examples in Table \ref{tab:cot-prompting-input-examples}.
We save both the final answer and the chain of thought for later analysis.

\section{Context-free pronoun predictions by model}
\label{sec:context-free-model-predictions-appendix}

\begin{figure*}[ht]
  \begin{tabular}{p{0.16\linewidth}p{0.8\linewidth}}
        \vspace{0.01mm} \small{he/him/his} & \includegraphics[valign=m,width=\linewidth]{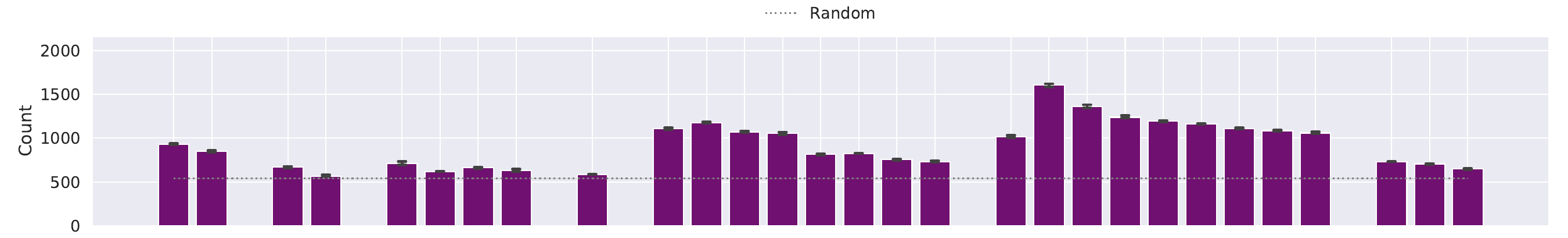} \\
        \vspace{-4mm} \small{she/her/her} & \includegraphics[valign=m,width=\linewidth]{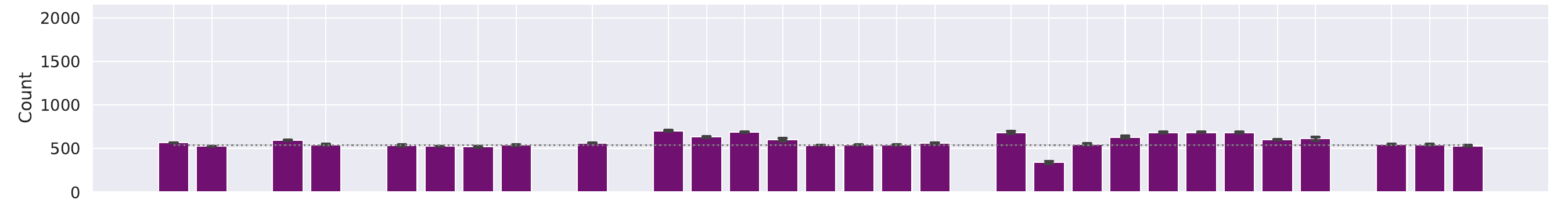} \\
        \vspace{-4mm} \small{they/them/their} & \includegraphics[valign=m,width=\linewidth]{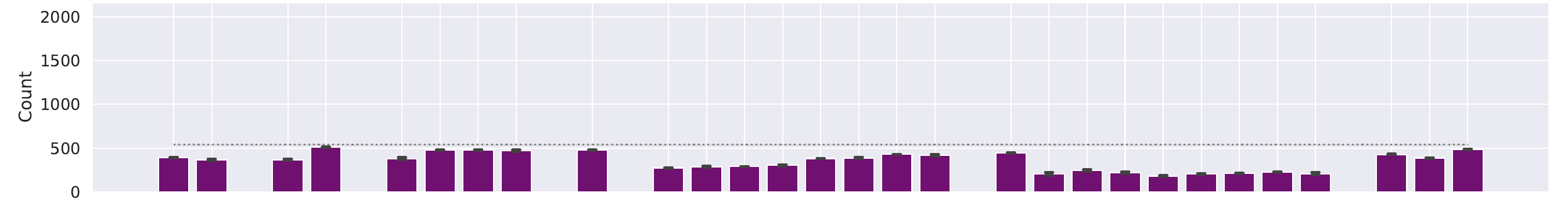} \\
        \vspace{-8mm} \small{xe/xem/xyr} & \includegraphics[valign=m,width=\linewidth]{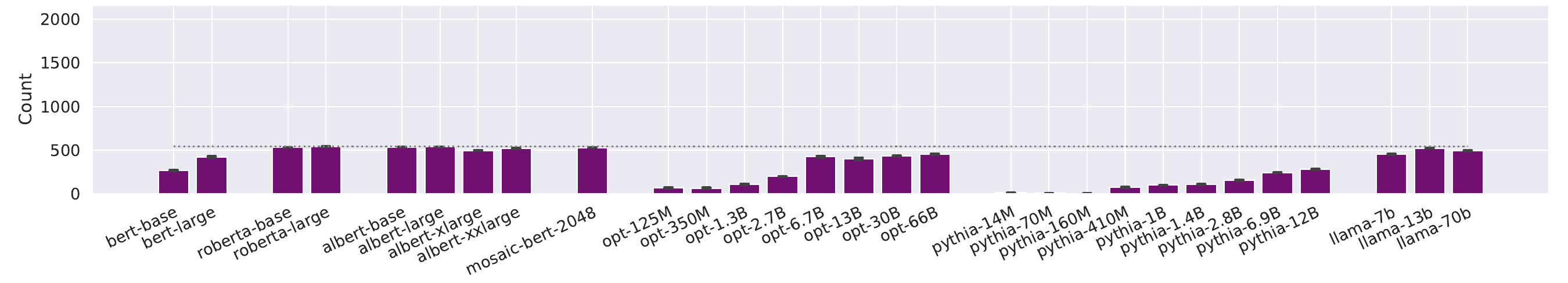} \\
    \end{tabular}
    \caption{Counts of pronoun predictions from all models, in the absence of context. The random baseline shows counts if each pronoun set was chosen equally often.}
    \label{fig:context-free-model-predictions-disaggregated}
    \vspace{-4mm}
\end{figure*}

Figure \ref{fig:context-free-model-predictions-disaggregated} shows per-model pronoun predictions in the absence of context. All models predict \textit{he/him/his} more frequently than \textit{she/her/her}, which is in turn predicted more frequently than \textit{they/them/their} and \textit{xe/xem/xyr}. However, encoder-only models are more balanced in their predictions across the four pronoun sets, compared to decoder-only models which show very stark differences in pronoun predictions.

\section{Results with Vanilla and Chain-of-Thought Prompting}
\label{sec:prompting-results}

\subsection{Vanilla prompting}
\begin{figure}[ht]
    \centering
    \begin{subfigure}[b]{0.49\linewidth}
        \centering
        \includegraphics[width=\linewidth]{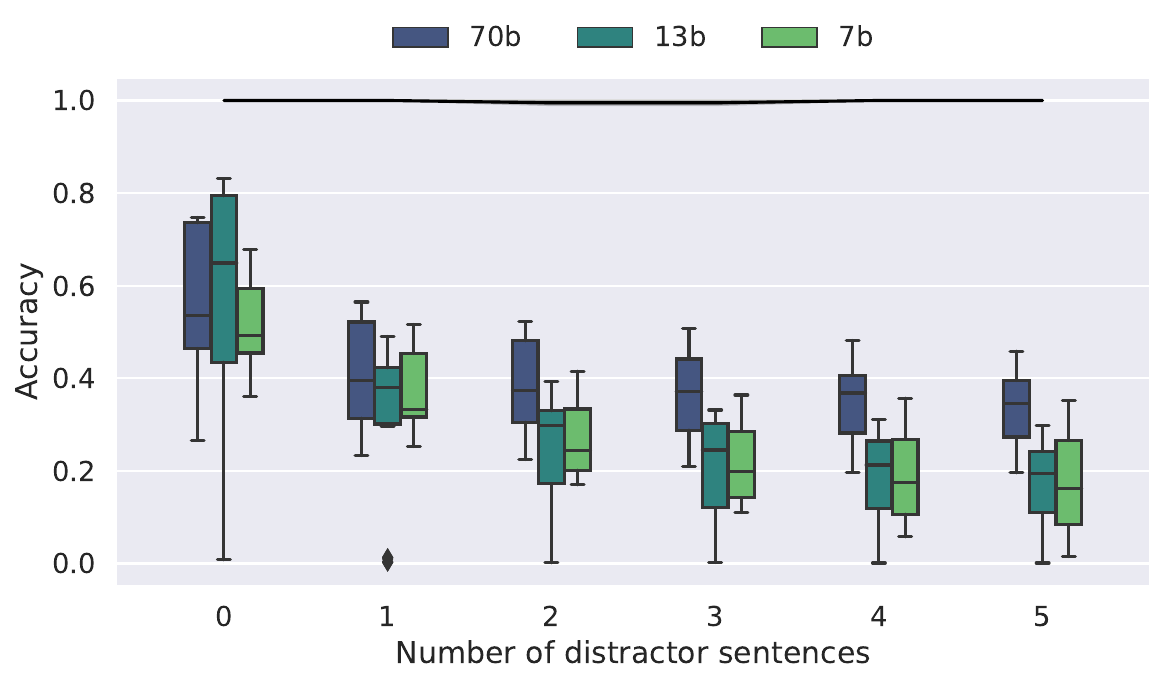}
        \caption{\textsc{Llama-2-chat}}
        \label{fig:llama-2-chat-prompting}
    \end{subfigure}
     \begin{subfigure}[b]{0.49\linewidth}
        \centering
        \includegraphics[width=\linewidth]{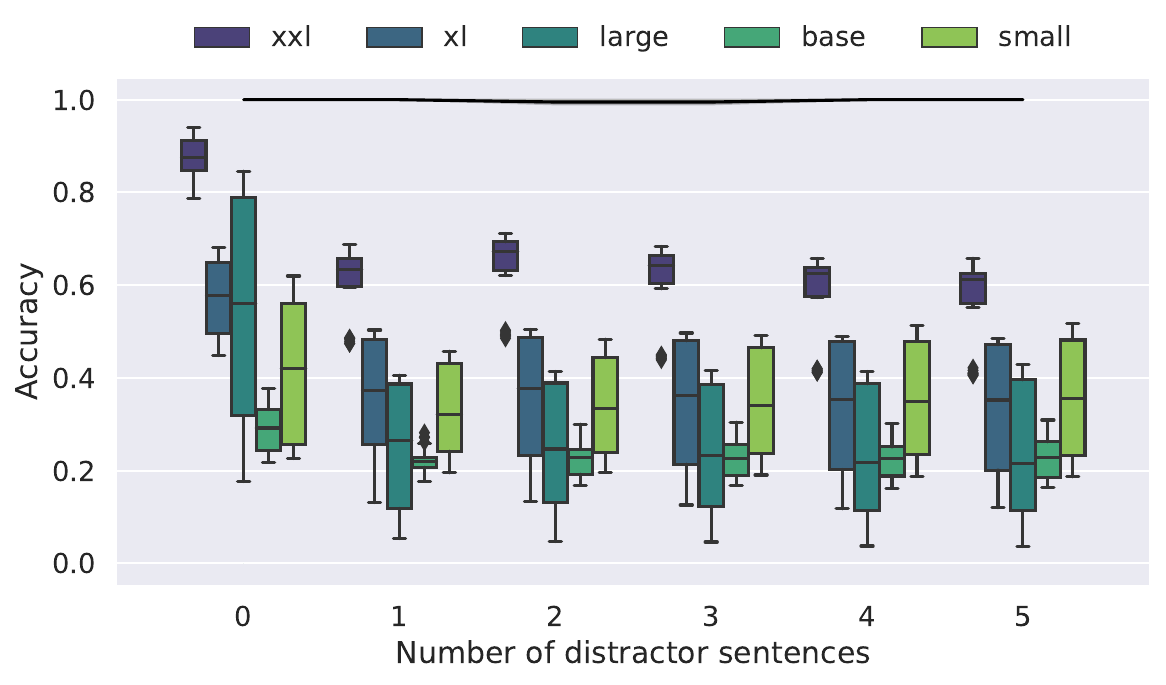}
        \caption{\textsc{FLAN-T5}}
        \label{fig:flan-t5-prompting}
    \end{subfigure}
    \caption{Performance of chat models (\textsc{Llama-2-chat} and \textsc{FLAN-T5}) with additional distractors, using vanilla prompting. The boxplots show the range of performance across 10 different templates.}
    \label{fig:prompting-scaling}
\end{figure}

Prompting is a different model evaluation mechanism than log likelihoods, with higher task demands that lead to lower performance than log likelihoods with both base models and instruction fine-tuned chat models~\citep{hu-levy-2023-prompting,hu2024auxiliary,kauf2024comparing}.
We thus expect vanilla prompting results (using the prompts listed in Appendix \ref{sec:prompting}) to be worse than results with log likelihoods.
Indeed, Figure \ref{fig:llama-2-chat-prompting} shows that \textsc{Llama-2-chat} prompting performance is lower than \textsc{Llama-2} evaluated with log likelihoods, even with no distractors.
Figure \ref{fig:flan-t5-prompting} shows the results of standard prompting with \textsc{FLAN-T5}, an encoder-decoder model which shows similar patterns of degradation to decoder-only models.
Bigger models are mostly better and degrade more gracefully than the smaller ones, but there remains a lot of variance across prompts, as shown in the box plots.

\subsection{Chain-of-thought prompting}

As \textsc{FLAN-T5-xxl} shows strong performance with low variance compared to all the other chat models we consider, we focus on this model for additional evaluation with chain-of-thought prompting.
Zero-shot chain-of-thought prompting encourages models to think step-by-step, which could in theory produce much better results on pronoun fidelity.
While chain-of-thought prompting is excessive for a task as simple as pronoun fidelity, it might encourage the model to explicitly list the referents and associated pronouns, which could help the model predict the correct pronoun with higher accuracy.
In practice, however, we find that it leads to worse performance, potentially due to hallucination.

\begin{figure}[ht]
    \centering
    \includegraphics[width=\linewidth]{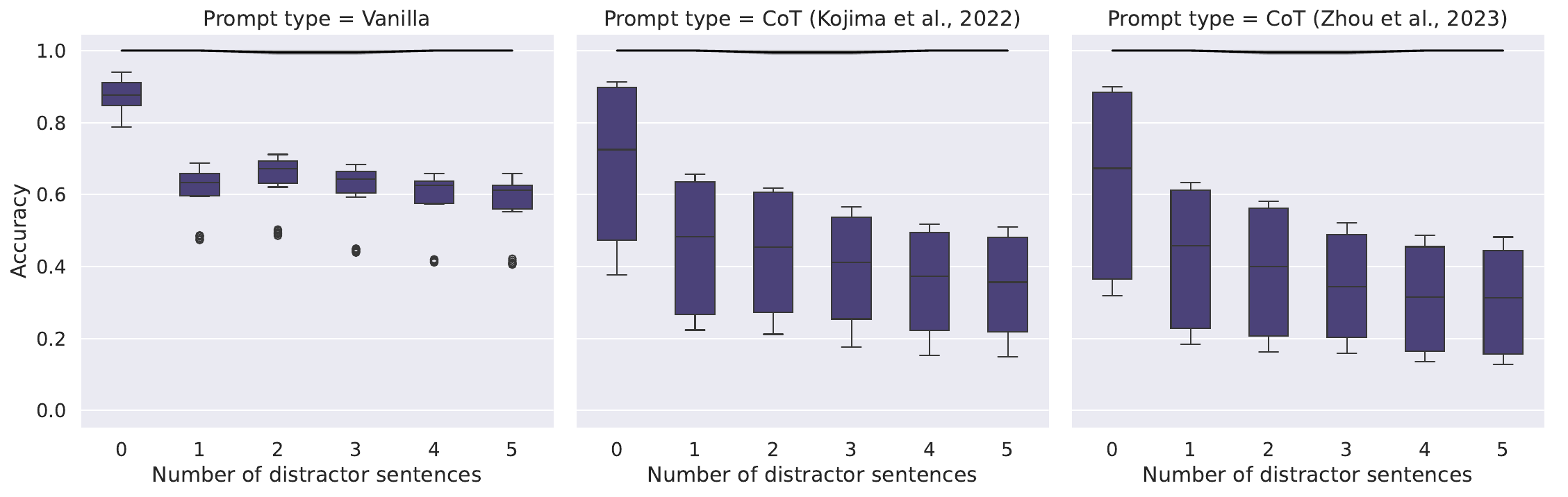}
    \caption{Performance of \textsc{FLAN-T5-xxl} with distractor sentences, comparing vanilla prompting to two types of chain-of-thought prompting. Here, the model's \textit{final answers} are used for evaluation and the boxplots show the range of performance across 10 different templates.}
    \label{fig:cot-prompting-answer}
\end{figure}

Figure \ref{fig:cot-prompting-answer} shows the pronoun fidelity of \textsc{FLAN-T5-xxl} with different types of prompting based on the final answer the model provides. Both types of chain-of-thought prompting worsen performance and increase the variance across prompts compared to vanilla prompting.

\begin{figure}[ht]
    \centering
    \includegraphics[width=\linewidth]{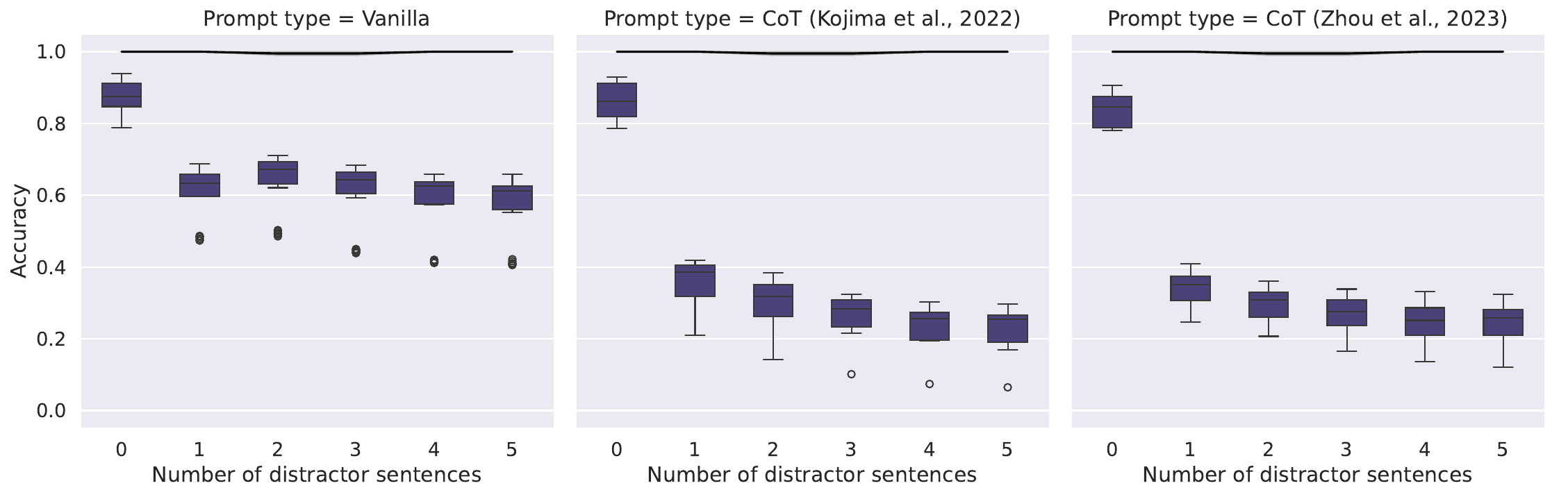}
    \caption{Performance of \textsc{FLAN-T5-xxl} with distractor sentences, comparing vanilla prompting to two types of chain-of-thought prompting. Here, the model's \textit{chain of thought} is used for evaluation and the boxplots show the range of performance across 10 different templates.}
    \label{fig:cot-prompting-cot}
\end{figure}

When examining model-generated answers and chains of thought, we found that \textsc{FLAN-T5-xxl} does not in fact solve the problem step by step as the instruction suggests.
Instead, the chain of thought often already contains an answer, and the final answer is not necessarily the same as this one.
Therefore, we also plot performance using answers from the model-generated chain of thought in Figure \ref{fig:cot-prompting-cot}. Once again, performance with 1-5 distractors is much lower, showing that chain-of-thought prompting degrades performance compared to vanilla prompting.
However, with no distractors, performance is almost exactly the same as vanilla prompting, as models simply generate the answer within the chain of thought.
This reinforces that chain-of-thought is unnecessary for a task this simple.

\end{document}